\documentclass{article} 

\pdfoutput=1

\PassOptionsToPackage{numbers, compress, sort}{natbib}

\usepackage{times}  
\usepackage{helvet}  
\usepackage{courier}  
\usepackage{url}  
\usepackage{graphicx}  
\usepackage{natbib}

\usepackage[utf8]{inputenc} 
\usepackage[T1]{fontenc}    
\usepackage{hyperref}       
\usepackage{url}            
\usepackage{booktabs}       
\usepackage{amsfonts}       
\usepackage{amsmath}  
\usepackage{amsthm}
\usepackage{amssymb}  
\usepackage{nicefrac}       
\usepackage{microtype}      
\usepackage{bbm}
\usepackage{xcolor}         
\usepackage{adjustbox}
\usepackage{geometry}
\usepackage{pdflscape}
\usepackage[skip=6pt plus 2pt, indent=12pt]{parskip} 

\usepackage{thmtools,thm-restate}

\usepackage{authblk}
\usepackage{algorithmicx}
\usepackage{algorithm}
\usepackage[noend]{algpseudocode}
\usepackage{mathtools}

\usepackage{subfigure}
\usepackage{tabularx}

\usepackage{basic}
\hypersetup{
  pdffitwindow=true,
  pdfstartview={FitH},
  pdfnewwindow=true,
  colorlinks,
  linktocpage=true,
  linkcolor=blue,
  urlcolor=blue,
  citecolor=blue
}

\newtheorem{proposition}{Proposition}

\newcommand{\E}[2]{\mathbb{E}_{#1}\left[#2\right]}

\newcommand{\Var}[2]{\mathrm{Var}_{#1}\left(#2\right)}
\newcommand{\Cov}[2]{\textrm{\textbf{Cov}}_{#1}\left[#2\right]}

\usepackage{xspace}

\newcommand{\X}[0]{\mathcal{X}}

\newcommand{\Y}[0]{\mathcal{Y}}
\newcommand{\lf}[0]{\lambda}

\newcommand{\R}[0]{\mathbb{R}}

\newif\ifsinglecolumn

\newcommand{\G}[0]{\mathcal{G}}

\newcommand{\name}[0]{\textsc{Smoothie}\xspace}
\newcommand{\namelocal}[0]{\textsc{Smoothie-Local}\xspace}
\newcommand{\nameglobal}[0]{\textsc{Smoothie-Global}\xspace}
\newcommand{\bestonval}[0]{\textsc{Best-on-Val}\xspace}
\newcommand{\random}[0]{\textsc{Random}\xspace}
\newcommand{\pairrm}[0]{\textsc{PairRM}\xspace}
\newcommand{\lknn}[0]{\textsc{Labeled-kNN}\xspace}
\newcommand{\bestensemble}[0]{\textsc{Best-Model}\xspace}

\newcommand{\distacc}[0]{\textsc{Distr-Acc}\xspace}
\newcommand{\distr}[0]{\textsc{Distr-Rouge2}\xspace}

\newcommand{\Dtrain}[0]{\mathcal{D}_{\text{train}}}

\newcommand{\Dtest}[0]{\mathcal{D}_{\text{test}}}

\newcommand{\ntrain}[0]{n_{\text{train}}}

\newcommand{\nn}[0]{\text{NN}}

\newcommand{\Vbar}{\bar{\mathcal{V}}}

\newcommand{\V}{\mathcal{V}}

\newcommand{\route}{\mathrm{route}}

\usepackage{etoolbox}

\makeatletter
\newcommand{\changeoperator}[1]{%
  \csletcs{#1@saved}{#1@}%
  \csdef{#1@}{\changed@operator{#1}}%
}
\newcommand{\changed@operator}[1]{%
  \mathop{%
    \mathchoice{\textstyle\csuse{#1@saved}}
               {\csuse{#1@saved}}
               {\csuse{#1@saved}}
               {\csuse{#1@saved}}%
  }%
}
\makeatother

\usepackage{enumitem}  

\setlist{} 

\setlist[itemize]{
    leftmargin=2em,
    itemsep=0.2em,
    parsep=0pt,
    topsep=3pt,
    partopsep=0pt
}

\setlist[enumerate]{
    leftmargin=2em,
    itemsep=0.2em,
    parsep=0pt,
    topsep=3pt,
    partopsep=0pt
}

\title{\name: Label Free Language Model Routing}

\author[1]{Neel~Guha$^*$}
\author[1]{Mayee~F.~Chen \thanks{Equal contribution. For correspondance, contact \url{nguha@stanford.edu} and \url{mfchen@stanford.edu}.}}
\author[1]{Trevor~Chow}
\author[1]{Ishan~S.~Khare}
\author[1]{Christopher~R\'e}

\affil[1]{Department of Computer Science, Stanford University}

\begin{document}

\maketitle

\begin{abstract}

    \noindent Large language models (LLMs) are increasingly used in applications where LLM inputs may span many different tasks. Recent work has found that the choice of LLM is consequential, and different LLMs may be good for different input samples. Prior approaches have thus explored how engineers might select an LLM to use for each sample (i.e. \textit{routing}). While existing routing methods mostly require training auxiliary models on human-annotated data, our work explores whether it is possible to perform \textit{unsupervised} routing. We propose \name, a weak supervision-inspired routing approach that requires no labeled data. Given a set of outputs from different LLMs, \name constructs a latent variable graphical model over embedding representations of observable LLM outputs and unknown ``true'' outputs. Using this graphical model, we estimate sample-dependent quality scores for each LLM, and route each sample to the LLM with the highest corresponding score. We find that \name's  LLM quality-scores correlate with ground-truth model quality (correctly identifying the optimal model on 9/14 tasks), and that \name outperforms baselines for routing by up to 10 points accuracy.

\end{abstract}
\section{Introduction}

Large language models (LLMs) are increasingly being deployed in \textit{multi-capability} regimes where data inputs may span a diverse range of tasks, each of which requires different capabilities~\cite{bommasani2021opportunities}. For instance, an LLM-powered chatbot may be asked to write code, answer questions about different domains, summarize documents, perform extraction, and more~\cite{arora2023language, chen2021evaluating, guha2024legalbench, bommasani2021opportunities}. One challenge is that while engineers often have access to numerous pre-trained LLMs (i.e., through Huggingface or various APIs), they do not know which LLM is optimal for each possible user input~\cite{shnitzer2023large}. Because the quality of generations can vary significantly between LLMs, choosing the right LLM for each input sample is important to ensure high task performance~\cite{jiang2023llmblender}.

Recent work has explored various ways to utilize ensembles of pretrained LLMs in multi-capability settings, by (1) collecting a diverse pool of LLMs and (2) identifying which LLM to \textit{route} each sample to~\cite{shnitzer2023large, lu2023routing}. However, most existing approaches require labeled data; Engineers typically either (1) train an auxiliary model using labeled data to rank or predict the LLM to which each sample should be routed~\cite{jiang2023llmblender, sakota_2024}, or (2) directly use labeled data to determine which LLM is the best on average~\cite{shnitzer2023large}. 
As a result, engineers designing routing protocols face the practical difficulty of constructing labeled datasets. 

Given a candidate pool of LLMs and an unlabeled test dataset, this paper explores how to best select LLM outputs for each sample in an entirely unsupervised manner---without labeled data, or models trained on labeled data. 
To make progress in addressing this question, we face two technical challenges:
\begin{itemize}
    \item \textbf{Unknown LLM quality}: The first challenge is estimating the quality of each LLM.
    Access to labeled data allows engineers to identify higher performing LLMs by measuring the accuracy/quality of LLM outputs. In this paper, we study the question of how to estimate quality \textit{without} labeled validation data.
    \item \textbf{Sample-conditional generator performance}: The second challenge is determining how to select the best LLM for each individual test sample. LLM outputs can vary in quality over different samples, which could render population-level estimates of LLM quality misleading.
\end{itemize}

\begin{figure}
    \centering
    \includegraphics[width=\linewidth]{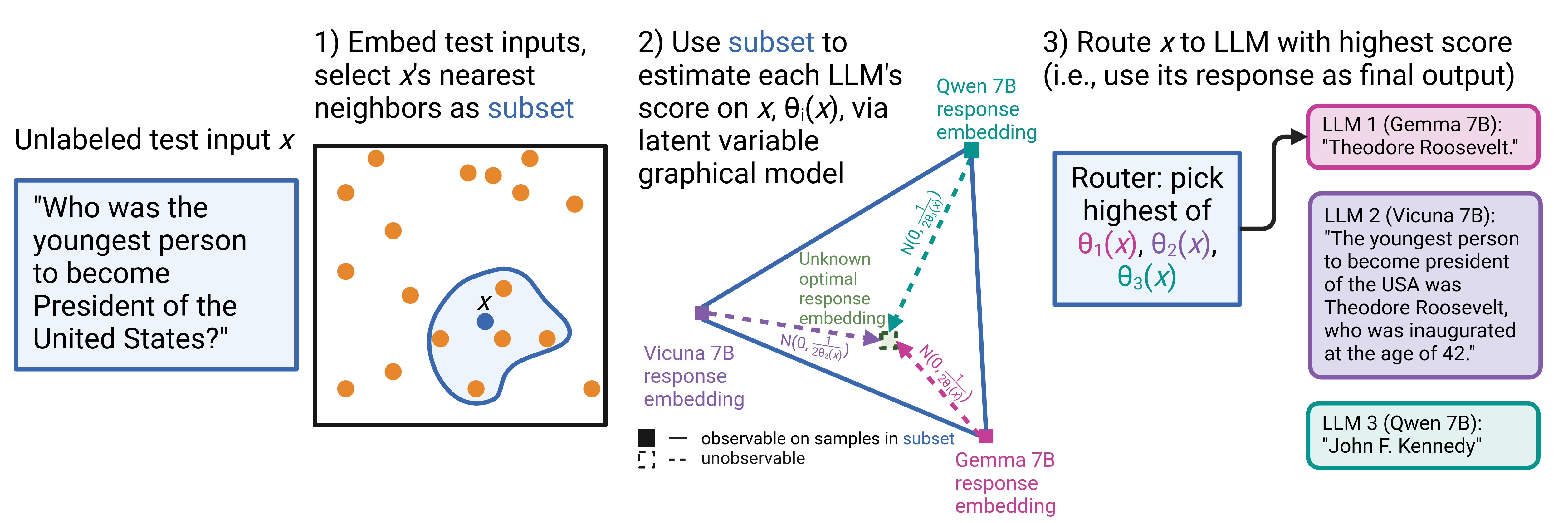}
    \caption{For a given input $x$, \name estimates the quality of every LLM ensemble's generation, and uses this quality weight to route $x$ to a single LLM.}
    \label{fig:banner}
\end{figure}

In this work, we propose \name, a method for routing samples to LLMs in a label-free manner (Figure~\ref{fig:banner}). Below, we describe how \name addresses the two challenges described above.
\begin{itemize}
    \item \textbf{Quality estimation}: Using the LLM outputs for each test sample as ``voters,'' \name estimates the quality of each generator using methods from Weak Supervision (WS). Concretely, \name constructs a latent variable graphical model over observable LLM outputs and an unknown true output. By modeling the embedding vector difference between each LLM output and the true output as a multivariate Gaussian, we can derive a closed-form estimator adapted from~\cite{shin2022universalizing} for learning LLM quality scores efficiently. 
    \item \textbf{Conditioning}: We condition these quality estimates to be particular to a given test sample by only using the nearest neighbors of a test sample in the training data as inputs to the estimator (i.e., kernel smoothing). We then route each test sample to the LLM with the highest quality score estimate on that sample. We call the version of \name that produces quality estimates using all available test data \nameglobal, and we call the version that uses a sample's nearest neighbors \namelocal.
\end{itemize}

We empirically evaluate \name in three stages.
\begin{itemize}
    \item \textbf{LLM selection:} First, we assess \nameglobal's ability to identify---from an ensemble of mixed quality LLMs---the optimal LLM for a given task overall. On traditional generation tasks such as summarization, reading comprehension, and data-to-text generation, we find that \nameglobal's learned LLM quality-weights correlate with actual LLM performance ($\rho = 0.72)$), and on the AlpacaEval benchmark, \nameglobal identifies the best-performing instruction model 70\% of the time~\cite{alpaca_eval}. The highest quality LLM identified by \nameglobal---all computed without labeled data---can beat random-selection by up to 15 points win-rate on AlpacaEval, and by up to 8 points on SQuAD.
    \item \textbf{Routing:} Second, we study whether \namelocal's sample-conditional scoring mechanism allows it to \textit{route} samples in mixed-task datasets to higher-performing LLMs (i.e., the multi-capability regime). We find that \namelocal can improve the quality of produced generations by up to 7 points accuracy over \nameglobal, and that \namelocal outperforms baseline unsupervised routing methods by up to 10 points accuracy and supervised routing methods by up to 5.0 points accuracy. 
    \item \textbf{Prompt selection:} Finally, we assess whether \name's quality-estimation mechanism can be applied to select the optimal prompt template in a candidate pool while using a fixed LLM. We find that \nameglobal can outperform other prompt selection approaches by up to 18 points, allowing a 410M parameter model to match the performance of 6.9B parameter model.
\end{itemize}

\section{Related Work}

We provide an abbreviated related work, with a full treatment in Appendix~\ref{app:rel_work}. 

\paragraph{Routing} Routing has been classically utilized in Mixture-of-Experts models~\cite{jordan1993hierarchical, jacobs1991adaptive, shazeer2017outrageously, fedus2022switch}, which involve jointly training a set of models as well as a router. Recently, routing mechanisms have been used at inference time to decide which pre-trained LLM to use for a given sample~\cite{sakota_2024}. Some approaches involve training an auxiliary model using labeled training data to either score or rank the performance of each LLM on each test sample~\cite{ jang2023exploring, ravaut2023summareranker}. Others do not involve training a model but instead use nearest neighbor methods, selecting the LLM that does the best on a test sample's labeled neighbors~\cite{lee2023orchestrallm, shnitzer2023large}. In contrast, \name does not require any labels.

\paragraph{Ensembling} Ensembling is another way of utilizing a pool of LLMs. Existing work has primarily focused on ensembling outputs for classification tasks~\cite{wang2022self, arora2022ask, pitis2023boosted}. Ensembling generative outputs typically requires training an auxiliary model~\cite{jiang2023llmblender}, combining or switching among outputs when decoding~\cite{izacard2020leveraging, shen2024learningdecodecollaborativelymultiple}, or averaging in weight space~\cite{wan2024fusechat}. 

\paragraph{Prompt selection} In addition to selecting the best LLM for a sample, prior works have studied how to select the best prompt or in-context examples. While the simplest approach is to use a held-out labeled dataset~\cite{perez2021true}, there are also retrieval-based approaches to selecting the best in-context examples~\cite{su2022selective}, as well as approaches based on mutual information~\cite{sorensen2022information} and probability-based measures~\cite{yang2023improving}, although the latter two are limited to classification.

\paragraph{Weak supervision} \name utilizes statistical techniques inspired by weak supervision, which programmatically generate labels for an unlabeled dataset by aggregating the predictions of several weak ``voters'' via a latent variable graphical model~\cite{ratner2017data, ratner2017snorkel}. Weak supervision has mostly been studied in classification settings~\cite{fu2020fast, ratner2018training} but more recently has been extended to tasks such as learning rankings and manifolds~\cite{shin2022universalizing, vishwakarma2022lifting}. We derive our estimation procedure from the Gaussian model in~\cite{shin2022universalizing}, applying it to LLM embeddings and the routing setting.

\section{Preliminaries}

\subsection{Problem setup}

Let $\V$ be the token vocabulary space, and let $\Vbar = \V \times \dots \times \V$ be the space of all vocabulary sequences. We consider a generative task with input text $x \in \X \subset \Vbar$ and reference output text $y \in \Y \subset \Vbar$.
We have a candidate pool of $m$ LLMs, $G = \{g_1, \dots, g_m \}$, where each $g_i \in \G: \X \rightarrow \Y$ produces a generative output sequence $g_i(x)$ for a given input text sequence $x$. 
We are given an unlabeled test dataset $\Dtest = \{x_i\}_{i = 1}^n$, where the ground-truth reference outputs are \textit{unknown.} 

Our goal is to route each sample $x \in \Dtest$ to one of the LLMs in $G$. Specifically, we wish to construct a router $\route: \G^m \times \X \rightarrow \G$ that selects the LLM that yields the highest quality generation on $x$ for each test sample $x$, without any labeled data. 

\subsection{Graphical model}

We present a probabilistic graphical model (see Figure~\ref{fig:banner} (center)) that describes how the LLM outputs, $g_1(x), \dots, g_m(x)$, are related to a true output $y$ in terms of each LLM's quality on a given input $x$, which we call $\theta_i(x)$, corresponding to each $g_i(x)$.
Let $z_{g_0}: \Vbar \rightarrow \R^d$ map from a sequence of tokens to a $d$-dimensional embedding using a common model $g_0$ such as SentenceBERT~\cite{reimers-2019-sentence-bert}. Define $\lf_i(x) := z_{g_0}([x, g_i(x)])$ to be the observable embedding of $x$ and the LLM output, and define $z^\star(x) := z_{g_0}([x, y])$ to be the latent ground-truth embedding of $x$ and reference output $y$. Similar to the approach in~\cite{shin2022universalizing}, we model the distribution over embedding vectors, $\Pr(z^\star(x), \lf_1(x), \dots, \lf_m(x) | x)$ as 
\begin{align}
    \Pr(z^\star(x), \lf_1(x), \dots, \lf_m(x) | x) = \frac{1}{Z} \exp \bigg(\sum_{i = 1}^m - \theta_i(x) \|\lf_i(x) - z^\star(x) \|^2\bigg) \label{eq:pgm}
\end{align}

where $Z$ is the log partition function and the $\theta_i(x)$s---the LLM quality scores---are canonical parameters of the graphical model. Intuitively, our model captures LLM quality by supposing that if $g_i$ is of high quality and $\theta_i(x)$ is very large, then it should be unlikely for the LLM output to be very different from the true output in terms of Euclidean distance in embedding space. Conversely, if $\theta_i(x)$ is small, we assign larger probability to the setting where $\lf_i(x)$ and $z^\star(x)$ differ significantly. Finally, note that this graphical model corresponds to a multivariate Gaussian. That is, the vector $[\lf_1(x) - z^\star(x), \dots, \lf_m(x) - z^\star(x)] \in \R^{d m}$ is Gaussian with mean $\mu = \vec{0}$ and a diagonal covariance matrix $\Sigma \in \R^{dm \times dm}$ with $\Sigma_{jj} = \frac{1}{2 \theta_{\lceil j / m \rceil}(x)} $. Intuitively, this means that the average difference vector between each $\lf_i$ and $z^\star(x)$ is centered, with its magnitude inversely proportional to the LLM score $\theta_i(x)$ and independent of other LLMs. Given this probabilistic graphical model, our goal is to learn each quality score $\theta_i(x)$ from the unlabeled test dataset and use these for improved routing.
\section{Method}

Given an unlabeled test dataset $\Dtest$ and a pool of LLMs $G$, \name consists of two steps:
\begin{enumerate}
    \item \textbf{Estimation}: The LLM quality scores $\theta_1(x), \dots, \theta_m(x)$ are learned for each $x \in \Dtest$ (Section~\ref{sec:ws}, Algorithm~\ref{alg:ws}).
    \item \textbf{Routing}: The LLM with the highest scores is selected, and its output is used as our final prediction for $x$ (Section~\ref{sec:select}). 
\end{enumerate}

We describe each step in the following sections.

\subsection{LLM score estimation} \label{sec:ws}

We describe how to estimate each $\theta_i(x)$s in the graphical model in~\eqref{eq:pgm} using only unlabeled data from $\Dtest$. Then, we describe how the LLM score estimate can be instantiated to be sample-conditional.

\begin{algorithm}[tb]
   \caption{\textsc{Estimate Scores}}
\begin{algorithmic}[1]
   \State {\bf Input:} unlabeled test dataset $\Dtest$, LLMs $G$, $n_0$ nearest neighbors parameter, $g_0$ embedding model with dimension $d$.
   \State For all $x \in \Dtest$ and $g_i \in G$, obtain the generator output $\vec{g}_i(x)$ and embed the input and generator output using model $g_0$ to get embedding $\lf_i(x) := z_{g_0}([x, g_i(x)])$.
   \For{$x \in \Dtest, g_i \in G$}
    \For{$j, k \neq i \in [m]$}
        \State Compute $\hat{\delta}_{ij}(x) = \frac{1}{n_0} \sum_{x' \in \nn_{n_0}(x)} \|\lf_i(x') - \lf_j(x')\|^2$, and similarly $\hat{\delta}_{ik}(x)$ and $\hat{\delta}_{jk}(x)$. 
        \State Set $\hat{\theta}_i^{jk}(x) = d / (\hat{\delta}_{ij}(x) + \hat{\delta}_{ik}(x) - \hat{\delta}_{jk}(x))$.
        \EndFor
    \State Compute averaged estimate $\hat{\theta}_i(x) = \frac{1}{\binom{m-1}{2}} \sum_{j, k \neq i} \hat{\theta}_i^{jk}(x)$.
   \EndFor
    \State \Return $\hat{\theta}_i(x)$ for all $x \in \Dtest, g_i \in G$.
\end{algorithmic}
\label{alg:ws}
\end{algorithm}

\paragraph{Computing $\theta_i(x)$}
Below, we state a simple property arising from the fact that~\eqref{eq:pgm} corresponds to a multivariate Gaussian with a diagonal covariance matrix.

\begin{proposition} \cite{shin2022universalizing}\label{prop:ws}
For any $i, j \in [m]$, it follows from the graphical model in~\eqref{eq:pgm} that
\begin{align}
 \E{}{\|\lambda_i(x) - \lambda_j(x) \|^2} = \E{}{\|\lambda_i(x) - z^\star(x) \|^2} + \E{}{\|\lambda_j(x) - z^\star(x) \|^2}.
\end{align}

\end{proposition}

The proof is in Appendix~\ref{app:ws_proof} and relies on the fact that off-diagonal entries of $\Sigma$ are $0$. Note that the left hand side of the equation is observable while the two expectations on the right are unknown. We can apply this equation to pairs of LLM embeddings over a triplet of $\lf_i, \lf_j, \lf_k$ to form a system of three equations with three unknown expectations. Solving, we have
\begin{align}\label{eq:smoothie_acc}
    \E{}{\|\lambda_i(x) - z^\star(x)\|^2} = \frac{1}{2} \big(\delta_{ij}(x) + \delta_{ik}(x) - \delta_{jk}(x) \big) \; \forall (i, j, k) \in [m],
\end{align}

where $\delta_{ij}(x) = \E{}{\|\lf_i(x) - \lf_j(x)\|^2}$. Since~\eqref{eq:pgm} is a multivariate Gaussian with $\Sigma_{jj} = \frac{1}{2\theta_{\lceil j/m\rceil}(x)}$, we can write $\theta_i(x)$ as the following function of $\E{}{\|\lambda_i(x) - z^\star(x)\|^2}$:
\begin{align}
\E{}{\|\lambda_i(x) - z^\star(x)\|^2} = \sum_{j = 1}^{d} \E{}{(\lambda_{i,j}(x) - z^\star_j(x))^2} = \sum_{j = 1}^{d} \Var{}{\lambda_{i,j}(x) - z^\star_j(x)} = \frac{d}{2\theta_i(x)},
\end{align}

where $\lf_{i,j}(x)$ and $z^\star_j(x)$ are the $j$th indices of the embeddings $\lf_i(x)$ and $z^\star(x)$ respectively.
Therefore, we can write $\theta_i^{jk}(x) = \frac{d}{\delta_{ij}(x) + \delta_{ik}(x) - \delta_{jk}(x)}$, where each $\delta_{ij}(x)$ can be estimated using the LLM outputs on $\Dtest$, and in practice in Algorithm~\ref{alg:ws} we estimate $\theta_i(x)$ by averaging $\theta_i^{jk}(x)$ over all ${m-1 \choose 2}$ pairs of $j, k \neq i$.

\paragraph{Sample-conditional estimation of $\theta_i(x)$} Note that the 
expectation in $\delta_{ij}(x) = \E{}{\|\lf_i(x) - \lf_j(x) \|}$ is over the randomness in $\lf_i(x), \lf_j(x)$ conditioned on a fixed point $x$. However, we only have one sample per $x$. One simple approach is to use the entire dataset to estimate $\theta_i(x)$, i.e., $\hat{\delta}_{ij}(x) = \frac{1}{n}\sum_{x' \in \Dtest} \| \lf_i(x') - \lf_j(x')\|^2$. We denote this as \nameglobal. However, in \nameglobal each $\theta_i(x)$ for $i \in [m]$ is a constant over the entire $\Dtest$. Therefore, we use nearest neighbor kernel smoothing to estimate each $\delta_{ij}(x)$ in a sample-dependent manner, an approach we call \namelocal. Concretely, for $x \in \Dtest$, define $\nn_{n_0}(x) \subset \Dtest$ as the $n_0 < n$ nearest neighbors of $x$ (excluding $x$ itself) in $f_0$'s embedding space. Then, we construct $\hat{\delta}_{ij}(x) = \frac{1}{n_0} \sum_{x' \in  \nn_{n_0}(x)} \|\lf_i(x') - \lf_j(x') \|^2$, and do the same for $\hat{\delta}_{ik}(x), \hat{\delta}_{jk}(x)$ to get a sample-conditional estimate of $\theta_i(x)$. The procedure for estimating $\theta_i(x)$ in \namelocal is outlined in Algorithm~\ref{alg:ws}.

\subsection{Routing}\label{sec:select}

Once we have estimates of $\hat{\theta}_i(x)$ for each of the $m$ generators by using Algorithm~\ref{alg:ws}, we can construct our $\route()$ function. We define $\route(\G, x) = g_i$ where $i = \arg\max \{\theta_1(x), \dots, \theta_m(x) \}$, which selects the highest scoring LLM for input $x$ based on $\hat{\theta}_i(x)$. We apply this on $\Dtest$ to determine the best LLM for each input sample.

\section{Results}\label{sec:results}
We empirically analyze \nameglobal and \namelocal, focusing on four questions: 
\begin{enumerate}
    \item How well does \nameglobal recover ground-truth LLM rankings over samples belonging to the same task (Section \ref{sec:results:llm_task_quality})?
    \item In multi-task datasets, how well can \namelocal perform unsupervised-routing, by identifying the best LLM for each sample (Section \ref{sec:results:routing})?
    \item Can \nameglobal and \namelocal be applied to select from or route between different prompts (Section \ref{sec:results:prompts})?
    \item How does \nameglobal and \namelocal's performance change as a function of different algorithmic choices (Section \ref{sec:results:ablations})?
\end{enumerate}

\subsection{Single-Task LLM Scoring}\label{sec:results:llm_task_quality}

\paragraph{Setup} We begin by evaluating whether \nameglobal can accurately learn the relative performance of different LLMs on a single task-dataset.
We study three categories of tasks. First, we consider 7 datasets corresponding to commonly-studied natural language generation (NLG) tasks~\cite{liang2023holistic}: CNN/DailyMail and XSum (summarization), SQuAD (reading comprehension), TriviaQA (factual recall), E2E and WebNLG (data-to-text generation), and LegalBench's Definition Extraction (text extraction)~\cite{moritz2015teaching, see-etal-2017-get, narayan2018dont, rajpurkar2016squad, arora2024simple, 2017arXivtriviaqa, novikova2017e2e, shimorina2018handling, web_nlg, guha2024legalbench}. We report Rouge2 for summarization and data-to-text generation tasks and accuracy for all others. For all tasks other than Definition Extraction we evaluate \nameglobal on a 1000 sample subset.\footnote{Definition Extraction has fewer than 1000 samples.} For these tasks, we consider two ensembles of LLMs at different size points. At the 3B size point, our ensemble consists of Pythia-2.8B~\cite{biderman2023pythia}, Gemma-2B~\cite{gemmateam2024gemma}, Incite-3B~\cite{together2023redpajama}, and Dolly-3B~\cite{DatabricksBlog2023DollyV2}. At the 7B size point, our ensemble consists of Llama-2~\cite{touvron2023llama}, Mistral~\cite{jiang2023mistral}, Vicuna~\cite{zheng2023judging}, Gemma-7B~\cite{gemmateam2024gemma}, and Nous Capybara~\cite{daniele2023amplifyinstruct}. We manually write a single prompt template for each task, and all model generations rely on this template.

Second, we consider two instruction-following benchmarks: AlpacaEval and MixInstruct~\cite{alpaca_eval, dubois2024length, dubois2023alpacafarm, jiang2023llmblender}. For AlpacaEval, we rely on responses accessible via the online leaderboard.\footnote{Responses are available on the AlpacaEval website: \url{https://tatsu-lab.github.io/alpaca_eval/}.} We identify 10 LLMs (each from a different base family), and download these models' responses to the AlpacaEval instructions. We conduct 10 different simulations, where in each simulation we randomly select 5 LLMs from our pool to function as an ensemble. Reported win-rates use the standard GPT-4 references. For MixInstruct, we use generations from an ensemble of 11 different LLMs originally studied in ~\cite{jiang2023llmblender}. Following \cite{jiang2023llmblender}, we measure generation quality using a ChatGPT-based rank. 

Finally, we consider a more ``reasoning-intensive'' task, GSM8K~\cite{cobbe2021training}. We consider an ensemble of three models: Gemma-7B, Phi-2~\cite{phi2}, and Llema-7b~\cite{azerbayev2023llemma}. We prompt each model to provide a chain-of-though reasoning~\cite{wei2022chain}, and apply \name to these generations. 

For all datasets, we apply \nameglobal using SentenceBERT (\texttt{all-mpnet-base-v2}) embeddings of generations~\cite{reimers-2019-sentence-bert}.

\paragraph{Results} We first measure how frequently the highest-weighted LLM according to \nameglobal corresponds to the best-performing LLM in the ensemble. We observe that \nameglobal selects the best-performing LLM for 4/7 tasks on the 3B ensemble, and for 5/7 tasks on the 7B ensemble (Figure \ref{fig:smoothie_nlg_ensemble_ranks}). On AlpacaEval, \nameglobal selects the best-performing LLM by win-rate for 8/10 ensembles, and the best performing LLM by length-controlled win-rate for 7/10 ensembles. On MixInstruct and GSM8K, \nameglobal again identifies the best-performing LLM in the ensemble.

\begin{figure}[h!]
    \subfigure[]{
        \includegraphics[width=0.3\textwidth]{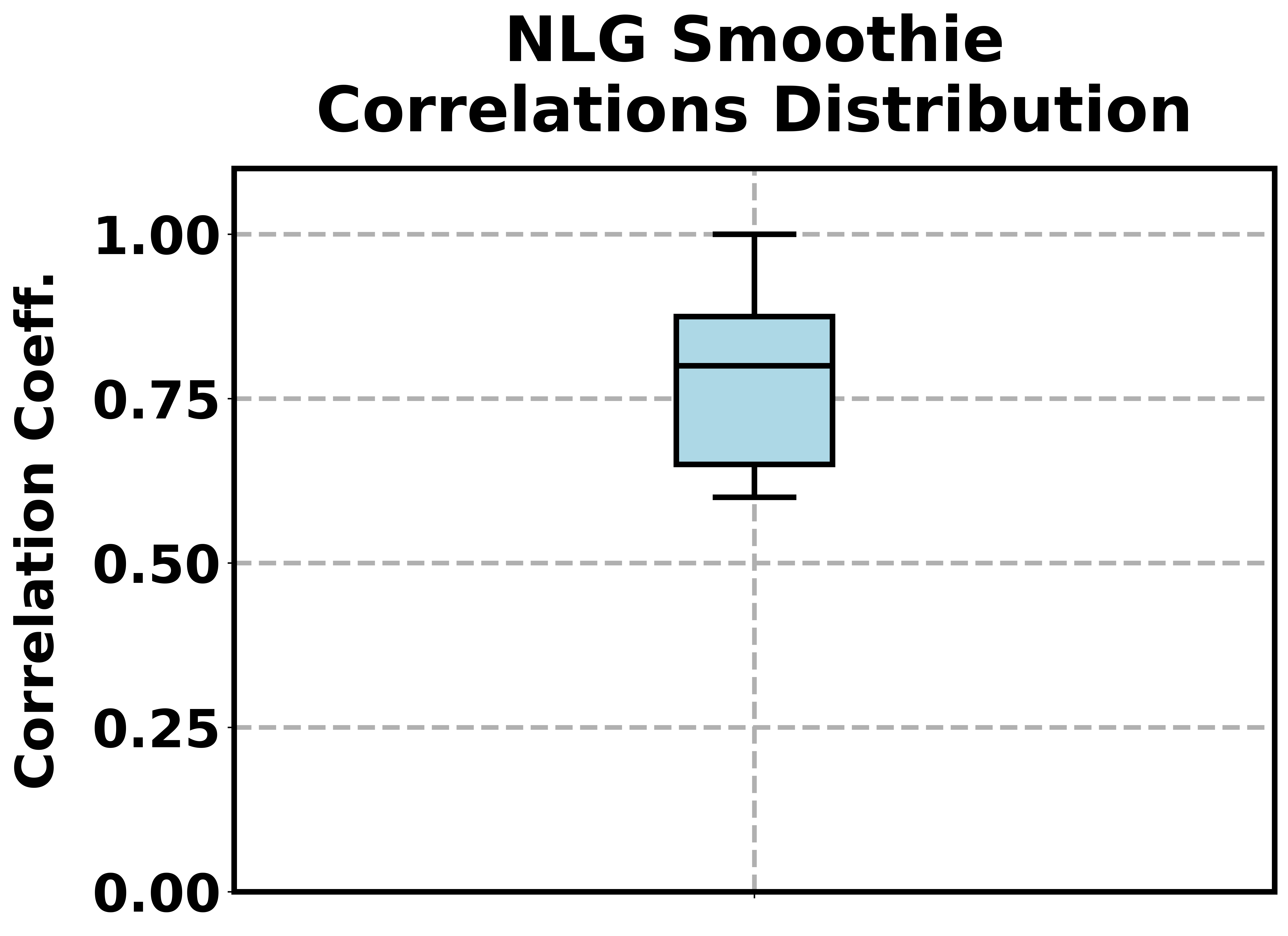}
        \label{fig:smoothie_weight_correlation}
    }
    \hfill
    \subfigure[]{
        \includegraphics[width=0.3\textwidth]{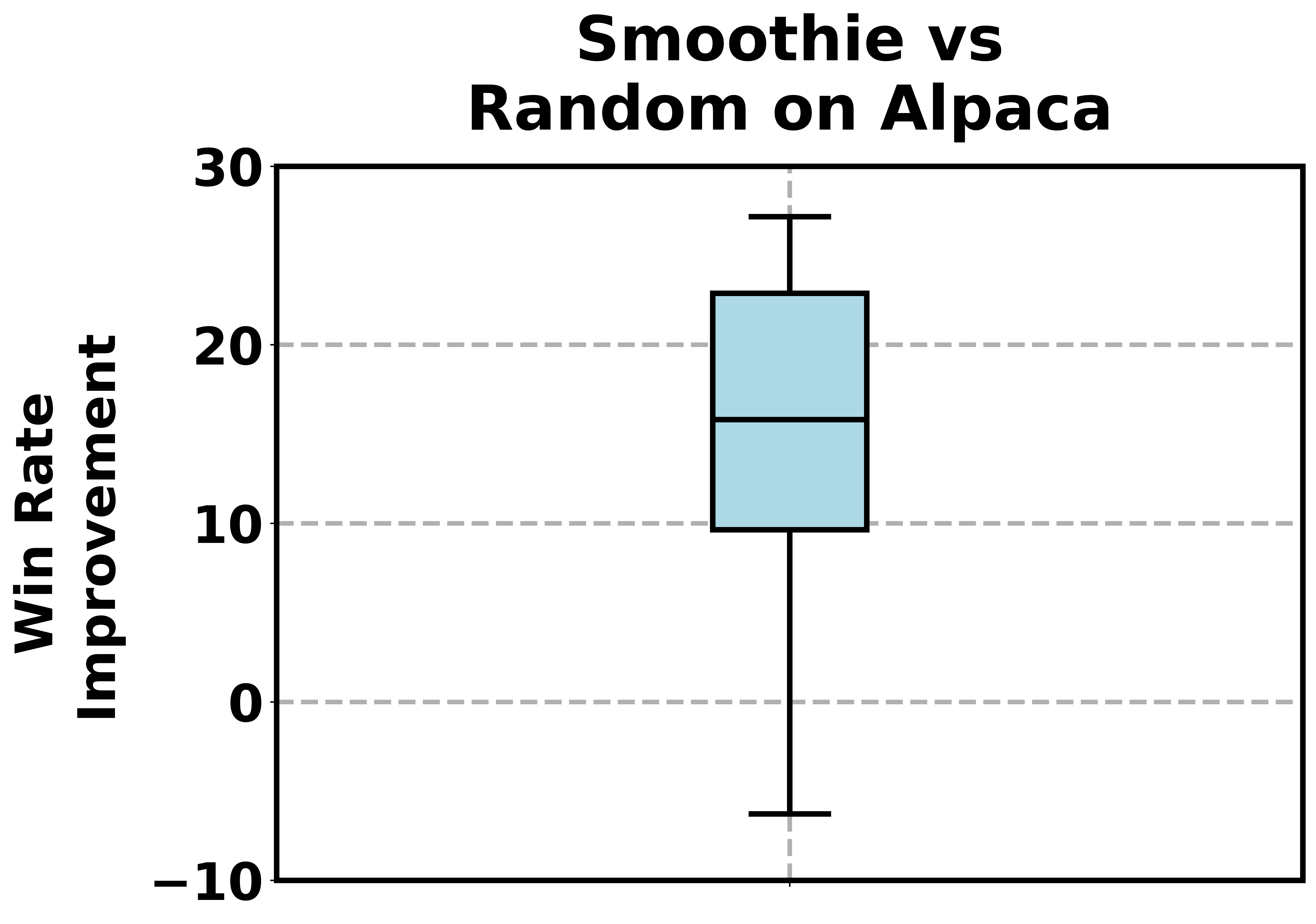}
        \label{fig:alpaca_win_rate_diff}
    }
    \hfill
    \subfigure[]{
        \includegraphics[width=0.3\textwidth]{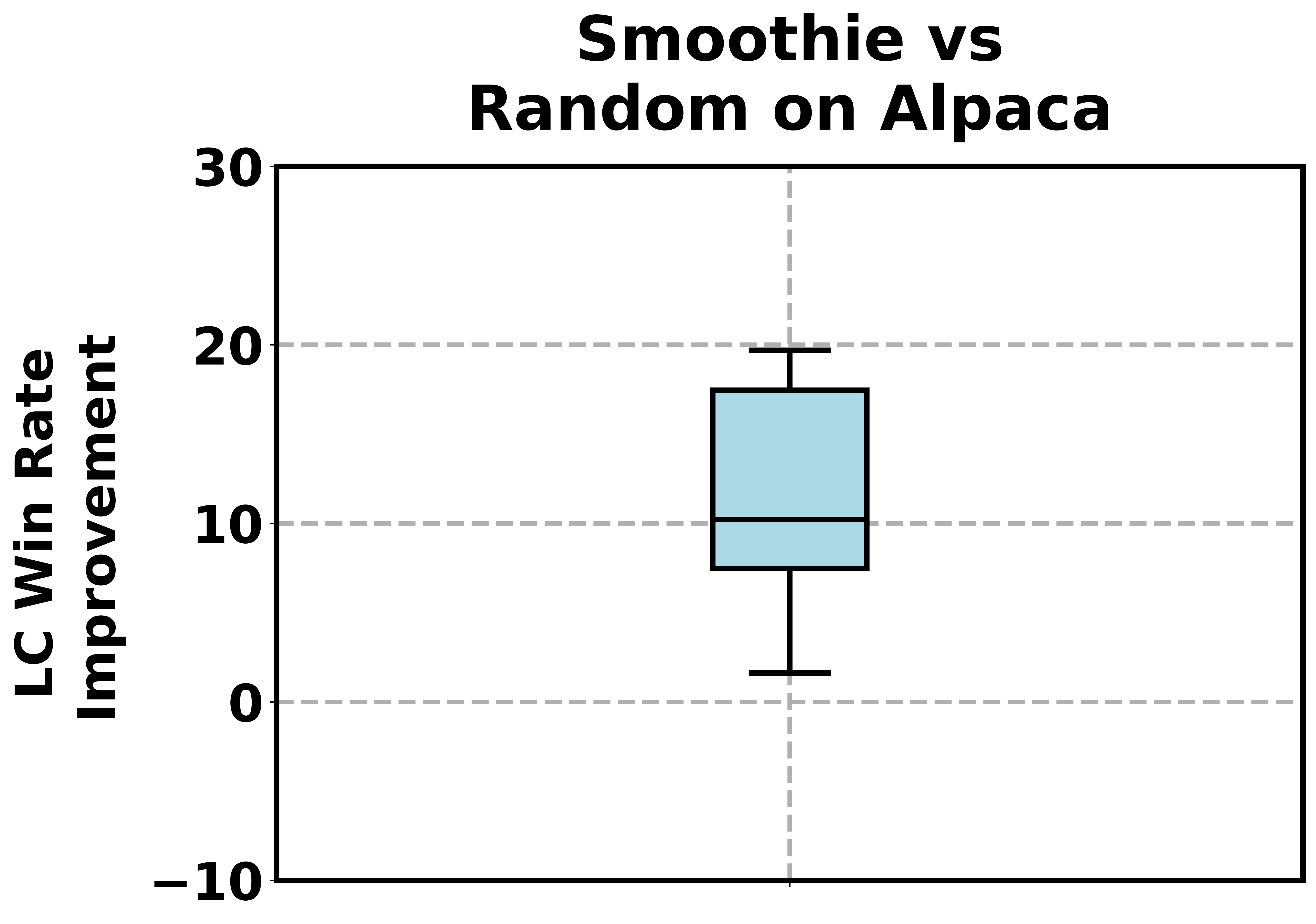}
        \label{fig:alpaca_lc_win_rate_diff}
    }
    
    \caption{\textbf{(a)} Spearman's rank correlation coefficient between \nameglobal weights and ground-truth LLM performance for 3B and 7B ensembles across NLG tasks. \textbf{(b)} \nameglobal's improvement over \random by win-rate on AlpacaEval. (c)\nameglobal's improvement over \random by length-controlled win-rate on AlpacaEval. }
    \label{fig:alpaca}
\end{figure}

Second, we measure how well \nameglobal captures quality differences between LLMs in the ensemble, by computing the Spearman's rank correlation coefficient between $\theta_i$ and ground truth quality scores ensemble models. Overall, we find that \nameglobal's learned weights approximate the relative ordering of model quality well. On the NLG tasks \nameglobal we measure an average correlation coefficient (across both ensembles and seven tasks) of 0.72. Figure \ref{fig:smoothie_weight_correlation} visually depicts the distribution of task coefficients---on only one ensemble/dataset pair is there a correlation coefficient $\leq 0$. On MixInstruct, we observe a correlation coefficient of 0.94, and on AlpacaEval, we observe a correlation coefficient of 0.46. 

Finally, we measure how the performance of the LLM selected by \name compares to other selection algorithms. We first compare \nameglobal to an unsupervised random baseline (\random), which would select a random model from the ensemble. We reported the \textit{expected} performance of this method, which is equivalent to taking the average performance of the ensemble. We also compare \nameglobal to a labeled baseline which simulates selecting an LLM on the basis of a small amount of validation data~\cite{perez2021true} (\bestonval). We sample a small labeled validation set (50 samples) and select the LLM that performs the best on this set. To account for sampling variation, we repeat this with 10 random draws and report the average performance. Because AlpacaEval has no training split and MixInstruct has no labeled data, we only compare \nameglobal to \random on those datasets.

\begin{table}[t]
\centering
\renewcommand{\arraystretch}{1.2}
\setlength{\tabcolsep}{7pt}
\begin{tabular}{@{}lcccccccc@{}}
\toprule
& & CNN & Def. Ext. & E2E & SQuAD & TriviaQA & WebNLG & XSum \\ \midrule
\multirow{3}{*}{3B}
& \small{\random} & 12.9 & 52.4 & 27.3 & 59.6 & \underline{32.7} & 23.4 & \underline{4.5} \\
& \small{\nameglobal} & \underline{\textbf{14.3}} & \underline{\textbf{61.5}} & \underline{\textbf{31.8}} & \underline{60.7} & 32.1 & \underline{\textbf{30.7}} & \underline{4.5} \\
& \small{\bestonval} & 13.0 & 60.5 & 31.1 & \textbf{66.4} & \textbf{38.7} & 30.3 & \textbf{5.3} \\
\midrule
\multirow{3}{*}{7B}
& \small{\random} & 13.7 & 58.5 & 35.3 & 67.9 & 59.3 & 44.1 & 6.9 \\
& \small{\nameglobal} & \underline{\textbf{14.5}} & \underline{\textbf{70.9}} & \underline{\textbf{36.9}} & \underline{\textbf{76.2}} & \underline{\textbf{68.3}} & \underline{45.9} & \underline{\textbf{8.4}} \\
& \small{\bestonval} & \textbf{14.5} & 69.4 & 36.7 & 74.0 & 65.8 & \textbf{48.3} & 8.3 \\
\bottomrule
\end{tabular}
\caption{Comparing \nameglobal to baseline methods on different ensembles across NLG datasets. Underlined values are the best performing \textit{unsupervised} methods. Bold values are the best performing \textit{overall} methods. We report rouge2 scores for CNN, XSum, WebNLG, and E2E, and accuracy for the rest. All metrics are scaled to 0-100.}
\label{tab:smoothie-global-comparison}
\end{table}

Table \ref{tab:smoothie-global-comparison} provides results for the seven NLG tasks. We find that \nameglobal outperforms the unsupervised \random baseline on 6/7 tasks for the 3B ensemble and on 7/7 tasks for the 7B ensemble. \nameglobal outperforms \random by up to 7pts (on tasks measured by rouge2), and by up to 12pts (on tasks measured by accuracy). We also observe that \nameglobal is frequently competitive with and even outperforms the \bestonval baseline, which uses labeled data. \nameglobal outperforms \bestonval on 4/7 tasks for the 3B ensemble, and 5/7 tasks for the 7B ensemble. On GSM8K, \nameglobal achieves a solve-rate of 37.5\% (matching \bestonval, while \random achieves a solve-rate of 28.3\% (Table \ref{tab:gsm8k_selection_results}).

\nameglobal also outperforms the \random baseline on the instruction-following datasets. On MixInstruct, \nameglobal achieves a ChatGPT-rank ($\downarrow$) of $3.91$, while \random achieves a ChatGPT-rank of $5.95$ (Table \ref{tab:mix_instruct_selection_results}). On AlpacaEval, \nameglobal outperforms \random on all but one trial (across both win-rate and length-controlled win-rate). \nameglobal outperforms \random by an average of 15pt win-rate, and up to 27pts. Figure \ref{fig:alpaca_win_rate_diff} and Figure \ref{fig:alpaca_lc_win_rate_diff} visualize this distribution.

\subsection{Multi-task Routing}\label{sec:results:routing}

\paragraph{Setup} We next assess whether \namelocal's sample-conditional scoring mechanism allows it to route samples to LLMs in the multi-capability regime. We construct two mixed-task distributions by combining existing datasets. The first distribution corresponds to tasks measured by accuracy, and contains SQuAD, TriviaQA, and Definition Extraction. We refer to this as \distacc. The second distribution corresponds to tasks measured by Rouge2, and contains CNN/DailyMail, XSum, Web NLG, and E2E. We refer to this as \distr. For each mixed-task dataset, we report the metric averaged across all tasks. We compare to three baselines. 

\begin{itemize}
    \item \random: A random-selection baseline which returns a generation from a random LLM in the ensemble. Though naive, prior work has found this to be a strong method in practice~\cite{lu2024blending}. We run 10 trials and report the mean of this approach to account for variance.
    \item \lknn: A labeled data-based KNN baseline. For this, we sample $50$ labeled samples from a separate hold-out set ($ \mathcal{D}_\text{val}$), and measure the performance of each candidate LLM on this set. For a given test sample $x$, we identify the $20$ most semantically similar instances in $ \mathcal{D}_\text{val}$ (using SentenceBERT embeddings~\cite{reimers-2019-sentence-bert}), and route $x$ to the highest performing LLM over this subset. We note that the \textsc{Labeled-KNN} baseline is derived from routing methods in \cite{shnitzer2023large, lee2023orchestrallm}. 
    \item \pairrm: A reward model from ~\cite{jiang2023llmblender} which accepts an instruction and multiple generations as input, scores each generations suitability for the instruction, and returns the predicted best generation. \pairrm is a labeled-data method which \cite{jiang2023llmblender} trained on collected preference data.
\end{itemize}

In addition, we also compare the best individual model in the ensemble (\bestensemble), and \nameglobal.  For both mixed-task datasets, we run \namelocal with SentenceBERT embeddings, and the sample-conditional version of \namelocal estimates $\theta_i(x)$ using a neighborhood size $n_0 = 1$.

\begin{table}[t]
\small
\centering
\renewcommand{\arraystretch}{1.3}
\setlength{\tabcolsep}{10pt}
\begin{tabular}{lccccc}
\toprule
& \multicolumn{2}{c}{\textbf{3B}} & \multicolumn{2}{c}{\textbf{7B}} \\
\cmidrule(lr){2-3} \cmidrule(l){4-5}
\textbf{Method} & \distacc & \distr & \distacc & \distr \\ \midrule
\small{\random} & 48.7 & 17.0 & 65.4 & 25.0 \\ \midrule
\small{\pairrm} & 53.9 & 19.0 & 71.8 & 25.5 \\ \midrule
\small{\lknn} & 51.0 & 16.8 & 71.7 & 26.2 \\ \midrule
\small{\bestensemble} & 52.3 & 18.1 & 73.2 & 26.4 \\ \midrule
\small{\nameglobal} & 51.3 & 18.1 & 66.5 & 26.1 \\ \midrule
\small{\namelocal} & \textbf{58.7} & \textbf{20.2} & \textbf{75.0} & \textbf{26.9} \\
\bottomrule
\end{tabular}
\caption{Comparing \namelocal to baseline methods on the 3B and 7B ensembles for multi-task distributions. \distacc and \distr are measured with accuracy and rouge2 respectively. Bold values indicate the best performing method for each dataset and model size. Metrics are scaled to 0-100.}
\label{tab:smoothie-local-comparison}
\end{table}

Results for the 3B and 7B ensembles over \distacc and \distr are provided in Table \ref{tab:smoothie-local-comparison}. We find that \namelocal outperforms all baselines across both data distributions, for both ensembles. Though \namelocal requires no labels, it still outperforms labeled data baselines like \lknn and \pairrm. We observe a substantial gap between \namelocal and \nameglobal, which indicates that \namelocal's sample-specific scoring mechanism provides performance improvements. 

\begin{figure}[h!]
    \centering
    \begin{subfigure}[]
        \centering
        \includegraphics[width=0.4\textwidth]{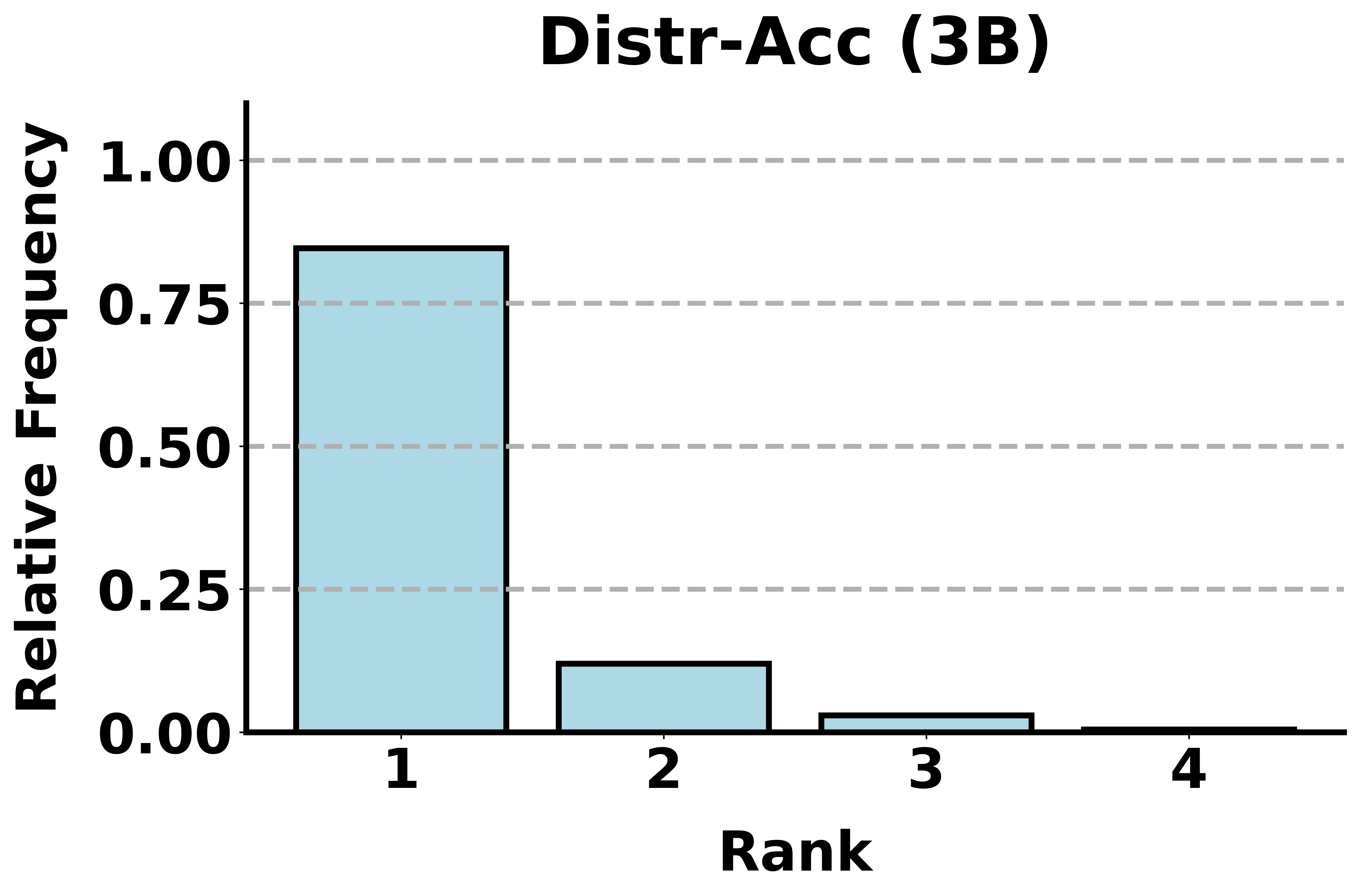}
        \label{fig:acc_group_3b_ensemble_rank_relative_frequency}
    \end{subfigure}
    \hfill
    \begin{subfigure}[]
        \centering
        \includegraphics[width=0.4\textwidth]{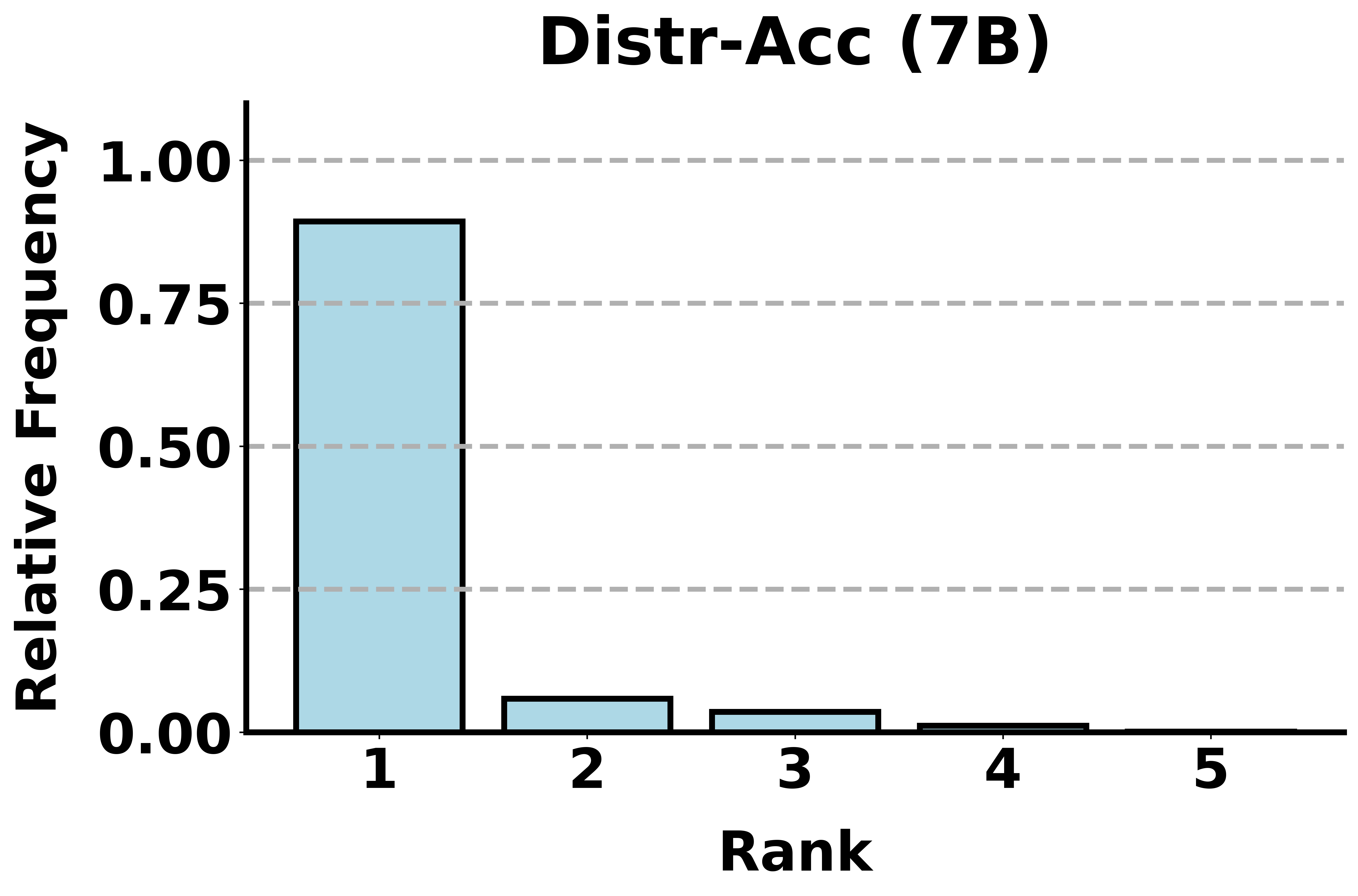}
        \label{fig:acc_group_7b_ensemble_rank_relative_frequency}
    \end{subfigure}
    \vskip\baselineskip
    \begin{subfigure}[]
        \centering
        \includegraphics[width=0.4\textwidth]{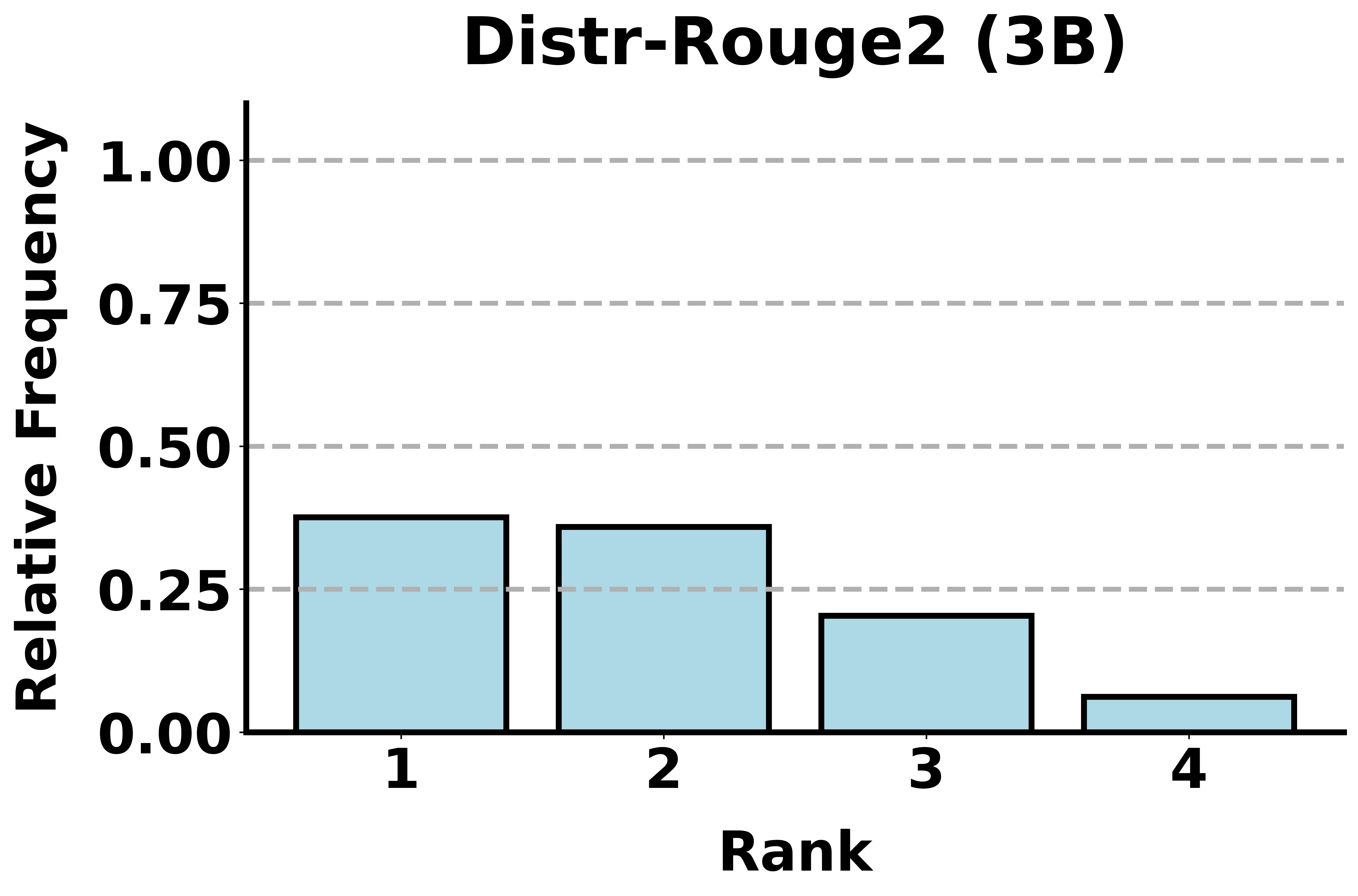}
        \label{fig:rouge2_group_3b_ensemble_rank_relative_frequency}
    \end{subfigure}
    \hfill
    \begin{subfigure}[]
        \centering
        \includegraphics[width=0.4\textwidth]{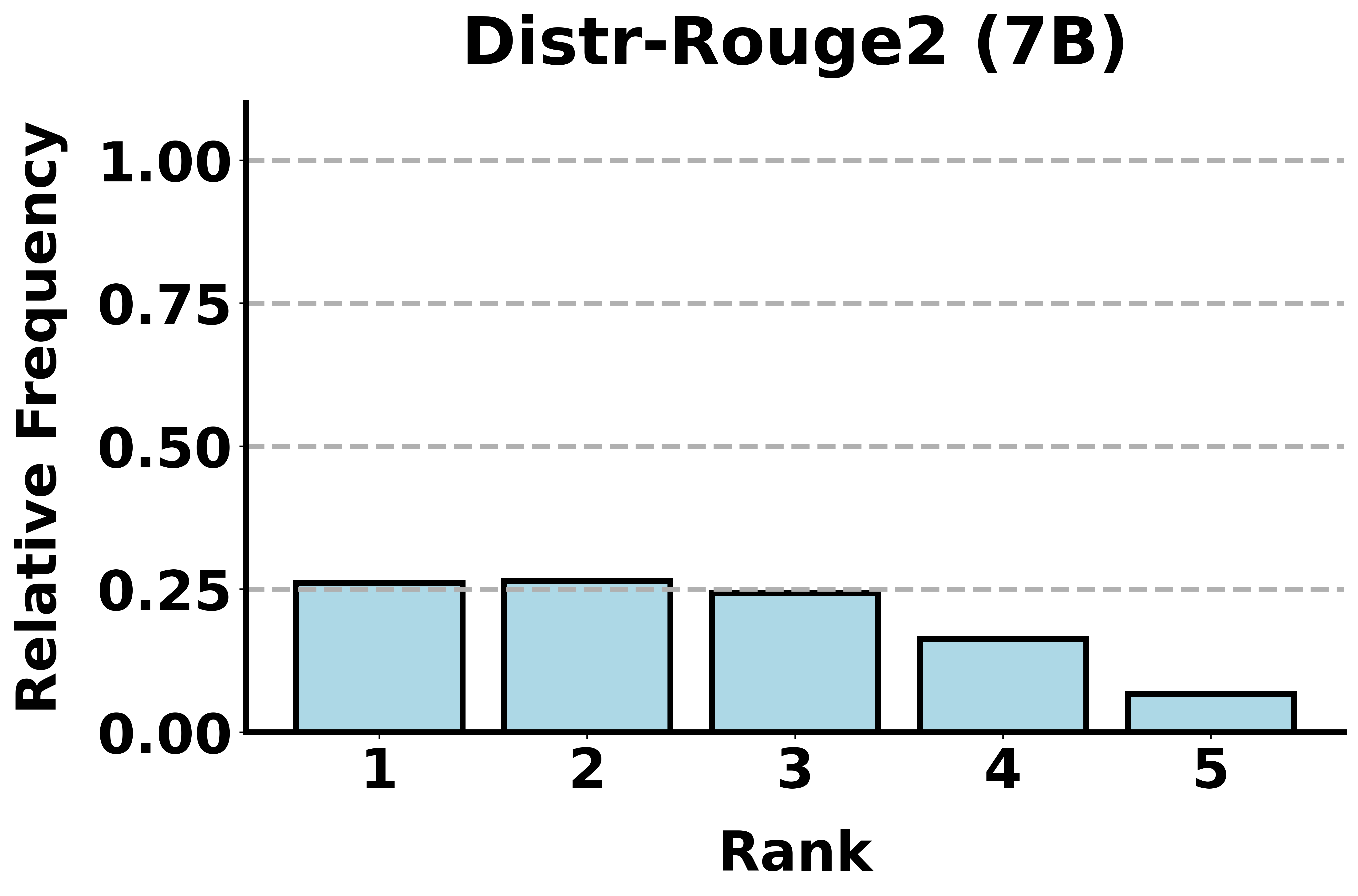}
        \label{fig:rouge2_group_7b_ensemble_rank_relative_frequency}
    \end{subfigure}
    \caption{On \distacc and \distr, we measure how frequently \namelocal selects the $i$-th best generation across the ensemble, for both the 3B and 7B ensembles. }
    \label{fig:smoothie_rank_distributions}
\end{figure}

Notably, we see that \namelocal substantially betters \bestensemble, indicating that \namelocal's routing mechanism is offering a performance improvement over a strategy which merely selects the best LLM on average. We study this in greater detail by examining the relative rank of the LLM selected by \namelocal for each sample. For each sample in \distacc and \distr, we rank the quality of each LLM's generation according to standard-competition ranking (i.e., ``1-2-2-4'' ranking). We then count how frequently \namelocal selects the rank-$i$ generation across each distribution for each ensemble. We visualize results in Figure \ref{fig:smoothie_rank_distributions}. As the visualizations demonstrate, \namelocal consistently selects the best or second-best generation from within the ensemble.

\subsection{Prompt Selection}\label{sec:results:prompts}

Third, we study whether \namelocal and \nameglobal  can be generalized to other settings where engineers have a candidate pool of text generators of unknown quality, and must select one of them to use for some application. In particular, we focus on the setting where an engineer has access to multiple prompt templates for a given generation task, and must select which prompt-templates' generation to use as the final output~\cite{guha2024embroid}. Unlike above, we assume the engineer only has access to one LLM. We study \namelocal and \nameglobal in this regime using the NLG tasks from Section \ref{sec:results:llm_task_quality}. For each task, we manually write between 3 and 5 prompt templates, varying the wording of instructions and the choice of in-context samples. We analyze \name applied to two models at different size points: Falcon (1B)~\cite{refinedweb} and Llama-2 (7B)~\cite{touvron2023llama}. 

Table \ref{tab:smoothie-prompt-comparison} provides the results. Overall, we find that \nameglobal selects the optimal prompt 2/7 times for Falcon-1B, and 3/7 times for Llama-2. \namelocal and \nameglobal consistently outperform \random--on 6/7 tasks for Falcon-1b and 6/7 tasks for Llama-2. On 7 task/model combinations, one of either \nameglobal or \namelocal matches or outperforms a labeled baseline. To better contextualize performance improvements from \nameglobal, we also compare to the improvement that accompanies increasing model size. Following a common practice in recent work, we can quantify the extent to which \nameglobal allows smaller models to match or exceed the performance of larger models~\cite{guha2024embroid, arora2022ask}. In Figure \ref{fig:scaling_results} (Appendix \ref{appendix:additional_results}), we compare \textsc{Random} and \nameglobal on models from the Pythia suite at four sizes: 410M, 1B, 2.8B, and 6.9B parameters~\cite{biderman2023pythia}. We observe that \nameglobal substantially improves performance---on E2E, \nameglobal enables a 410M parameter model to outperform a 6.9B parameter model.

\begin{table}[t]
\centering
\renewcommand{\arraystretch}{1.2}
\setlength{\tabcolsep}{7pt}
\begin{tabular}{@{}lcccccccc@{}}
\toprule
& & \small{CNN} & \small{Def. Ext.} & \small{E2E} & \small{SQuAD} & \small{TriviaQA} & \small{WebNLG} & \small{XSum} \\ \midrule
\multirow{3}{*}{\small{Falcon}}
& \small{\random} & \small{7.1} & \small{60.3} & \small{27.8} & \small{47.3} & \small{22.0} & \small{\underline{29.2}} & \small{4.7} \\
& \small{\nameglobal} & \small{\underline{7.9}} & \small{\underline{62.2}} & \small{\underline{\textbf{31.6}}} & \small{\underline{\textbf{53.3}}} & \small{\underline{\textbf{31.4}}} & \small{28.3} & \small{\underline{6.4}} \\
& \small{\namelocal} & \small{8.0} & \small{\textbf{69.2}} & \small{31.5} & \small{\textbf{53.3}} & \small{27.4} & \small{30.8} & \small{6.0} \\
& \small{\textsc{Best-on-Val}} & \small{\textbf{8.4}} & \small{64.2} & \small{31.0} & \small{52.7} & \small{\textbf{31.4}} & \small{\textbf{32.5}} & \small{\textbf{6.7}} \\
\midrule
\multirow{3}{*}{\small{Llama-2}}
& \small{\random} & \small{\underline{7.3}} & \small{47.8} & \small{31.6} & \small{54.0} & \small{45.9} & \small{45.5} & \small{11.2} \\
& \small{\nameglobal} & \small{6.9} & \small{\underline{\textbf{64.6}}} & \small{\underline{\textbf{37.6}}} & \small{\underline{61.4}} & \small{\underline{\textbf{68.7}}} & \small{\underline{48.5}} & \small{\underline{12.8}} \\
& \small{\namelocal} & \small{9.5} & \small{59.3} & \small{33.6} & \small{63.1} & \small{61.3} & \small{48.0} & \small{12.7} \\
& \small{\textsc{Best-on-Val}} & \small{\textbf{11.8}} & \small{\textbf{64.6}} & \small{35.0} & \small{\textbf{66.1}} & \small{\textbf{68.7}} & \small{\textbf{48.7}} & \small{\textbf{13.0}} \\
\bottomrule
\end{tabular}
\caption{Comparing \nameglobal and \namelocal to baseline methods in the prompt-selection setting. Underlined values are the best performing \textit{unsupervised} methods. Bold values are the best performing \textit{overall} methods. We report rouge2 scores for CNN, XSum, WebNLG, and E2E, and accuracy for the rest. All metrics are scaled to 0-100.}
\label{tab:smoothie-prompt-comparison}
\end{table}

\subsection{Ablations}\label{sec:results:ablations}

Finally, we conduct ablations to examine different aspects of \nameglobal and \namelocal: improving its efficiency, adjusting the neighborhood size, varying the choice of embedding model, and using different LLM ensembles.

\textbf{Improving efficiency}\; First, we explain \name's current efficiency properties. To estimate the Smoothie weights for routing, we use a simple closed-form procedure that does not require any SGD or training, as described in Algorithm~\ref{alg:ws}. As a result, \name weights on the entire dataset can be computed in seconds---for the 7B ensemble, \namelocal on the multi-task datasets takes 2.14 seconds per 1000 samples, and \nameglobal on the single-task datasets takes under 0.03 seconds per 1000 samples. Moreover, \name does not require any ground-truth annotations; however, all $m$ model generations per test sample are needed as input to the algorithm. That is, we need $n \times m$ generations for a $\Dtrain$ of size $n$ samples.

Fortunately, the need for computing all model generations per test sample can be removed with a small algorithm tweak, making Smoothie even more efficient and its runtime independent of $n$. Suppose we have a held-out set of $\ntrain$ train samples with precomputed generations from the models in the ensemble. For each test sample, we retrieve the most similar train samples, learn the Smoothie weights for the sample using the corresponding train sample generations, and return the model with the highest Smoothie weight (i.e., in line 5 in Algorithm~\ref{alg:ws}, KNN is now over a held-out training dataset). This approach, which we call \name-\textsc{Train}, selects the model for a test sample without needing model generations for that sample. Only $\ntrain \times m$ generations are needed, regardless of how large the test dataset $n$ is.

We study the NLG tasks, using $\ntrain = 250$ samples. In Table \ref{tab:smoothie-train-test-selection-comparison} (Appendix \ref{appendix:additional_results}), we evaluate a version of \nameglobal-\textsc{Train}) and observe that it matches \nameglobal on 12/14 model-dataset pairs, and performs worse on the remaining 2/14 pairs. We also evaluate \namelocal-\textsc{Train}, on \distacc and \distr (Table \ref{tab:smoothie-train-test-local-comparison}) using a neighborhood of size $n_0 = 20$. We find here that while \namelocal-\textsc{Train} underperforms \namelocal on both the 3B and 7B ensemble for both \distacc and \distr, it still outperforms \random and remains competitive with supervised baselines.

\textbf{Neighborhood size}\; We study the impact of $n_0$, and consider \namelocal's performance for $n_0 \in [1, 5, 10, 20, 50, 100]$. Figure \ref{fig:smoothie_n0_ablation_acc_group} provides performance over \distacc and Figure \ref{fig:smoothie_n0_ablation_rouge2_group} provides performance over \distr. Overall, we find that \namelocal's performance steadily degrades as $n_0$ increases, and is highest when $n_0 = 1$.

\textbf{Choice of embeddings} \;We study how the choice of embeddings affects \nameglobal performance (Table \ref{tab:smoothie-routing-embedding-ablation}). Specifically, we compare the performance of \namelocal using Sentence-Bert embeddings (\texttt{all-mpnet-base-v2})~\cite{reimers-2019-sentence-bert} to BGE embeddings (\texttt{bge-small-en-v1.5})~\cite{bge_embedding}. We observe that \namelocal appears robust to different embeddings---\namelocal with BGE embeddings still outperforms other labeled and unlabeled baselines. Interestingly, we observe that certain embedding models appear to yield better performance over certain distribution/ensemble combinations. For instance, \namelocal with SentenceBERT embeddings outperforms \namelocal with BGE embeddings on \distacc for the 3B ensemble and \distr for the 7B ensemble, while performing worse on \distr for the 3B ensemble and \distacc for the 7B ensemble.

\textbf{Different ensembles}\; Finally, we consider whether \nameglobal can generalize to a wider array of ensembles (Figure \ref{fig:smoothie_many_ensemble_ablation}). We combine the LLMs contained in the 3B and 7B ensembles into a single pool, and sample $50$ distinct ensembles ranging in size from 4-7 LLMs. For each of the $7$ NLG tasks, we evaluate \nameglobal's ability to identify the best model from within each ensemble. Across these $350$ settings, we find that \nameglobal identifies the best model in $211$ of them ($60.2$\% of the time), and one of the two best models in 292 of them ($83$\% of the time).

\section{Conclusion}
In this paper we study and propose an algorithm for learning label-free routers for generative tasks. We validate our approach across a variety of evaluation regimes, finding it consistently beats other unsupervised approaches and often matches/exceeds supervised approaches.

\textbf{Limitations} We discuss several of \name's limitations. First, its multivariate Gaussian graphical model currently uses a diagonal covariance matrix. This assumes independent error vectors for each generation, though \name could be extended to account for dependencies~\cite{ratner2018training, varma2019learningdependencystructuresweak}. Additionally, \name optimizes only for performance without considering cost tradeoffs between large and small models. Finally, its reliance on embeddings may capture only certain aspects of semantic similarity. Other embedding models and additional heuristics could be used to create richer input features for \name.
\section{Acknowledgements}

We thank Gautam Machiraju, Jon Saad-Falcon, Krista Opsahl-Ong, Sabri Eyuboglu, Jordan Juravsky, Vishnu Sarukkai, Ben Spector, Eric Nguyen, Jerry Liu, Chris Fifty, Avanika Narayan, Michael Zhang, Vincent Chen, and Fred Sala for their helpful feedback and discussion. This research project has benefitted from the Microsoft Accelerate Foundation Models Research (AFMR) grant program.

We gratefully acknowledge the support of NIH under No. U54EB020405 (Mobilize), NSF under Nos. CCF2247015 (Hardware-Aware), CCF1763315 (Beyond Sparsity), CCF1563078 (Volume to Velocity), and 1937301 (RTML); US DEVCOM ARL under Nos. W911NF-23-2-0184 (Long-context) and W911NF-21-2-0251 (Interactive Human-AI Teaming); ONR under Nos. N000142312633 (Deep Signal Processing); Stanford HAI under No. 247183; NXP, Xilinx, LETI-CEA, Intel, IBM, Microsoft, NEC, Toshiba, TSMC, ARM, Hitachi, BASF, Accenture, Ericsson, Qualcomm, Analog Devices, Google Cloud, Salesforce, Total, the HAI-GCP Cloud Credits for Research program,  the Stanford Data Science Initiative (SDSI), and members of the Stanford DAWN project: Meta, Google, and VMWare. NG is suported by the Stanford Interdisciplinary Graduate Fellowship (SIGF). The U.S. Government is authorized to reproduce and distribute reprints for Governmental purposes notwithstanding any copyright notation thereon. Any opinions, findings, and conclusions or recommendations expressed in this material are those of the authors and do not necessarily reflect the views, policies, or endorsements, either expressed or implied, of NIH, ONR, or the U.S. Government.

\newpage

\bibliography{main}
\bibliographystyle{plain}

\newpage
\appendix
\section{Appendix}\label{appendix:front_matter}

In Appendix~\ref{appendix:glossary}, we provide a glossary of notation used in the paper. In Appendix~\ref{app:rel_work}, we provide an extended related work, and in Appendix~\ref{app:ws_proof} we provide a proof of Proposition~\ref{prop:ws}, which is used in deriving the \name algorithm. Finally, in Appendix~\ref{appendix:additional_results} we provide additional experimental results and details.

Code for reproducing our results and using \name is available at \url{https://github.com/HazyResearch/smoothie}.
\section{Notation}\label{appendix:glossary}
The glossary is given in Table~\ref{table:glossary} below.

\begin{table*}[hp!]
\centering
\begin{tabular}{l l}
\toprule
Symbol & Used for \\
\midrule
$\Vbar$ & The space of all vocabulary sequences. \\
$x$ & Input text $x \in \X \subset \Vbar$. \\
$y$ & Reference output text $y \in \Y \subset \Vbar$. \\
$G$ & Candidate pool of $m$ LLMs, $G = \{g_1, \dots, g_m\}$, where each $g_i \in \G: \X \rightarrow \Y$ \\
& produces a generation $g_i(x)$ on input $x$. \\
$\Dtest$ & Unlabeled test dataset $\Dtest = \{x_i\}_{i=1}^n$.  \\
$\route$ & Routing function $\route: \G^m \times \X \rightarrow \G$ that selects the best LLM from $G$ for each sample.\\
$\theta_i(x)$ & Quality score of the $i$th LLM on test sample $x$, also used in the graphical model in~\eqref{eq:pgm}. \\
$z_{g_0}$ & Embedding mapping $z_{g_0}: \Vbar \rightarrow \R^d$ for any text sequence, where $g_0$ is an embedding model \\
& such as SentenceBERT~\cite{reimers-2019-sentence-bert}. \\
$\lambda_i(x)$ & The observable embedding of $x$ concatenated with the $i$th LLM's generated output, \\
& $\lambda_i(x) := z_{g_0}([x, g_i(x)])$. \\
$z^\star(x)$ & The latent embedding of $x$ concatenated with unknown reference output, $z^\star(x) := z_{g_0}([x, y])$.  \\
$Z$ & Partition function for normalization of~\eqref{eq:pgm}. \\ 
$n_0$ & Number of nearest neighbors used to learn $\theta_i(x)$ for $x$. $n_0 = n$ (i.e., the entire test dataset)\\
& corresponds to \nameglobal and $n_0 < n$ corresponds to \namelocal. \\
$\hat{\delta}_{ij}(x)$ & The average squared Euclidean distance between the $i$th and $j$th LLM embeddings over \\
& a neighborhood around $x$, $\hat{\delta}_{ij}(x) = \frac{1}{n_0} \sum_{x' \in \nn_{n_0}(x)} \|\lf_i(x') - \lf_j(x')\|^2$. \\
& This is the primary expression used in computing $\theta_i(x)$. \\
\toprule
\end{tabular}
\caption{
	Glossary of variables and symbols used in this paper.
}
\label{table:glossary}
\end{table*}

\section{Extended Related Work}\label{app:rel_work}

\textbf{LLM Routing} The problem of determining how to route samples to various models has been long studied in statistics~\cite{jordan1993hierarchical, jacobs1991adaptive} as well as Mixture of Experts deep neural networks~\cite{fedus2022switch, shazeer2017outrageously}. These works focus on how to jointly train the models and router in a stable and efficient manner. 

Since many LLMs are now available off-the-shelf, recent works study how routing mechanisms can be applied at inference time to trained models. 
Some works involve training task or domain-specific expert models and then learning a router. The router can be a nearest neighbors algorithm~\cite{jang2023exploring}, a neural network~\cite{wang2024fusing} that classifies among the different domains corresponding to the experts, or an extra gate learned when training the expert models~\cite{muqeeth2024learningroutespecializedexperts}. These approaches do not explicitly require labels, but they require knowledge of what domain is used to train each expert and assume that each expert is the best model for its corresponding domain, therefore effectively using this mapping as a form of labels. In contrast, our setting focuses on routing among pre-trained LLMs where we do not know what models are optimal on what tasks and their samples.

A second category of inference-time routing works studies how to choose among a collection of pre-trained LLMs, which is the setting that \name focuses on. Several approaches involve training a meta-model that either scores or ranks how a LLM will perform on a sample~\cite{sakota_2024,jiang2023llmblender,ravaut2023summareranker}, all of which required labeled data to train. MoRE~\cite{si2023gettingmixturelanguagemodel} involves training a simpler random forest classifier, using the rate of agreement among LLMs as one of the features, which is similar to how Smoothie estimates scores; however, it also requires labeled data to train the classifier.  Some approaches~\cite{lee2023orchestrallm, shnitzer2023large} do not require training routers and simply use nearest neighbor methods. However, these nearest neighbor methods still use labeled data to determine what training samples each LLM performs the best on. \cite{lu2023routing} invokes a trained reward model for the routing mechanism. \cite{ding2024hybridllmcostefficientqualityaware} trains a classification-based router using the BARTScore metric on LLM generations as pseudolabels; this avoids using manually labeled data, demonstrating that while a majority of routing methods require labeled data, there exist some alternatives that do not. We leave it to future work to compare and integrate \name with other unsupervised approaches. 

Finally, complementary to our setting are works that jointly focus on cost minimization as well as quality of generations. RouterBench~\cite{hu2024routerbench} creates a benchmark for studying the cost-quality tradeoffs in routing systems. Optimizing for cost can be done algorithmically, such as in FrugalGPT~\cite{chen2023frugalgpt}, AutoMix~\cite{madaan2024automix}, RouteLLM~\cite{ong2024routellmlearningroutellms}, and~\cite{singla2023biobjectiveepsilonconstrainedframeworkqualitycost}, as well as via hardware enhancements such as SambaNova Systems' Composition of Experts~\cite{prabhakar2024sambanova}.

\textbf{LLM Ensembling} A rich literature has observed that ensembling LLM outputs---across different prompts or base models---can improve the accuracy of generated predictions. Prior work has proposed and studied a number of different ensembling algorithms for classification tasks, including majority-voting~\cite{liu2023pre, wang2022self}, weak-supervision~\cite{guha2024embroid, arora2022ask}, boosting~\cite{pitis2023boosted, hou2023promptboosting, zhang2024prefer}, and others~\cite{peng2022model, schick2020exploiting, lester2021power}.

More relevant to our work here is a literature on ensembling for generative tasks. One category of methods rely on an auxilliary sequence-to-sequence models to ``fuse'' generations from different prompts or base LLMs~\cite{jiang2023llmblender}. Though recently applied in the context of modern LLMs, the concept of fusion traces back to older work on summarization~\cite{barzilay-mckeown-2005-sentence, ravaut2023summary, lebanoff2020learning, lebanoff2020cascade}. Some techniques combine or switch among multiple outputs at inference time~\cite{izacard2020leveraging, huang2024enabling, shen2024learningdecodecollaborativelymultiple, mavromatis2024packllmsmodelfusion, wang2024mixtureofagentsenhanceslargelanguage}, while others involve averaging in weight space~\cite{jang2023exploring, ilharco2023editing, wan2024fusechat}. Lastly, ensembling can also be approximated by randomly selecting a model to be used in multi-turn settings~\cite{lu2024blending}.

\textbf{Other LLM Selection Algorithms} Beyond the setting of selecting among multiple LLMs, other works have explored how to select the optimal prompt template from a collection of candidate prompts. These works can be grouped into two categories. The first category assumes that engineers have access to labeled data. In the naive case, this labeled data can simply be used to select the best performing prompt~\cite{perez2021true, kumar2021reordering, rubin2021learning}. Another subset of this category focuses on the setting where new prompts can be generated by selecting in-context demonstrations from a set of labeled samples (typically a small training set)~\cite{do2024automatic, liu2021makes}. Prior work has proposed different methods for identifying the optimal in-context demonstrations to use, depending on the sample for which the LLM is being used to produce a prediction for~\cite{zhang2022automatic, rubin2021learning, su2022selective, wu2022self, zhang2022active, chang2022data}. The second category focuses on zero-label prompt selection methods, but solely for classification tasks~\cite{liao2022zero, sorensen2022information, yang2023improving}. Prior work here selects prompts on the basis of mutual information~\cite{sorensen2022information}, agreement rates between predictions produced by different prompts~\cite{liao2022zero}, and various probability based measures~\cite{yang2023improving, lu2021fantastically, gonen2022demystifying}.

\textbf{Weak supervision} \name utilizes techniques inspired by weak supervision literature. Weak supervision aims to programmatically generate labels on an unlabeled dataset by aggregating the predictions of several weak ``voters'', such as heuristics, pre-trained models, and knowledge graphs~\cite{ratner2017data, ratner2017snorkel}. It assumes a particular latent variable graphical model and uses its structure to estimate latent quantities, such as the accuracy of each voter (in our setting, the quality score of each LLM). Typically, this graphical model is a binary Ising model, as weak supervision has generally been studied in classification settings~\cite{ratner2018training, fu2020fast}, where embeddings have been utilized as auxiliary signal but not modeled explicitly~\cite{guha2024embroid, chen2022shoring}. Weak supervision has been applied to broader settings, such as for learning rankings, graphs, and manifolds~\cite{shin2022universalizing, vishwakarma2022lifting}. We derive our estimation procedure from the Gaussian model in~\cite{shin2022universalizing}, applying it to LLM embeddings. While both \name and~\cite{shin2022universalizing} use a multivariate Gaussian model, in \name we apply it to model routing with SBERT embeddings on natural language datasets, whereas~\cite{shin2022universalizing} conducts synthetic experiments in hyperbolic spaces and metric spaces induced by synthetic graphs. Moreover, \name uses nearest neighbor kernel smoothing to allow for sample-dependent weights---critical for routing---while~\cite{shin2022universalizing} calculates one global set of weights over the dataset. 

\textbf{Consistency-based selection} Consistency is central to unsupervised selection and aggregation methods, the simplest being majority vote. While weak supervision methods~\cite{fu2020fast} and \name heavily rely on notions of voter agreement as depicted in a graphical model, there are several other consistency-based methods. Minimum Bayes Risk methods~\cite{kobayashi-2018-frustratingly, bertsch2023itsmbrwaydown} selects the generation that has the highest average similarity (i.e., cosine) with other generations. This is similar to \name, which routes to the lowest value of~\eqref{eq:smoothie_acc}. If we ignore the subtraction of $\delta_{jk}(x)$ in~\eqref{eq:smoothie_acc} and average over more than just $\delta_{ij}(x)$ and $\delta_{ik}(x)$, then \name with $n_0 = 1$ is equivalent to~\cite{kobayashi-2018-frustratingly} Therefore, \name can be considered as a slightly modified and more general version of this approach. Another approach~\cite{CHAI20221029} relies on consistency between a ``global'' and ``local'' embedding for each generation. They solve an optimization problem that estimates each generation’s quality score by constructing a loss that enforces that the similarity between the estimated true generation (produced by a weighted average of candidates) and the candidate generation should be the same according to both global and local embeddings.
In contrast, \name uses one embedding space, relies on a multivariate Gaussian structure among embeddings, and does not require gradient descent to learn the quality of each generation.

\textbf{Test-Time Compute} Approaches like model routing, ensembling, and selection can all be seen as ways of utilizing \textit{test-time compute} to produce higher-quality generations from a system of LLMs. Test-time compute can also be utilized over a single LLM via techniques such as those used in OpenAI's o1, Chain of Thought, and Rephrase and Respond~\cite{openaio1, wei2023chainofthoughtpromptingelicitsreasoning, deng2024rephraserespondletlarge}. Other works have recently studied how test-time compute scales~\cite{brown2024largelanguagemonkeysscaling, chen2024llmcallsneedscaling}---finding that producing more generations can often yield the correct response---and how to combine multiple test-time methods, such as Archon~\cite{saadfalcon2024archonarchitecturesearchframework}. It is interesting future work to consider how \name can be integrated with other test-time compute techniques.
\section{Proof of Proposition~\ref{prop:ws}}\label{app:ws_proof}

We provide a proof of proposition~\ref{prop:ws}, which is a direct property of multivariate Gaussians that is also presented in~\cite{shin2022universalizing}. We first expand $\E{}{\|\lf_i(x) - \lf_j(x) \|^2}$:
\begin{align}
    &\E{}{\|\lf_i(x) - \lf_j(x) \|^2} = \E{}{\|(\lf_i(x) - z^\star(x)) - (\lf_j(x) - z^\star(x)) \|^2} \label{eq:decomposition}\\
    &= \E{}{\|\lf_i(x) - z^\star(x) \|^2} + \E{}{\|\lf_j(x) - z^\star(x) \|^2} - 2\E{}{(\lf_i(x) - z^\star(x))^\top (\lf_j(x) - z^\star(x))} \nonumber 
\end{align}

Let $\lf_{i,k}(x)$ be the $k$th element of the $\lf_i(x)$ embedding, and similarly define $z^\star_k(x)$. Note that since $\Sigma$ is diagonal, we can write 
\begin{align*}
&\Cov{}{\lf_{i,k}(x) - z_k^\star(x), \lf_{j,k}(x) - z_k^\star(x)} \\
&= \E{}{(\lf_{i,k}(x) - z_k^\star(x) )\cdot (\lf_{j,k}(x) - z_k^\star(x))} - \E{}{\lf_{i,k}(x) - z_k^\star(x)} \E{}{\lf_{j,k}(x) - z_k^\star(x)} \\
&=0
\end{align*}
for all $k \in [d]$. Since $\mu = \vec{0}$, we thus have that $\E{}{(\lf_{i,k}(x) - z_k^\star(x) )\cdot (\lf_{j,k}(x) - z_k^\star(x))} = 0$ for all $k \in [d]$, which implies that $\E{}{(\lf_i(x) - z^\star(x))^\top (\lf_j(x) - z^\star(x))} = 0$. Plugging this into~\eqref{eq:decomposition}, we have
\begin{align}
 \E{}{\|\lambda_i(x) - \lambda_j(x) \|^2} = \E{}{\|\lambda_i(x) - z^\star(x) \|^2} + \E{}{\|\lambda_j(x) - z^\star(x) \|^2}.
\end{align}
\section{Additional Experiments and Details}\label{appendix:additional_results}
This section contains additional details on experiments discussed in Section \ref{sec:results}.

\subsection{Datasets and Models}

Table \ref{tab:huggingface_datasets_links} provides links to the Huggingface datasets used for each task. For E2E, CNN/DailyMail, XSum, and Web NLG we measure performance using rouge2. For SQuAD, TriviaQA, and Definition Extraction we measure using ``accuracy.'' A model generation is treated as ``correct'' if if contains the answer, and incorrect otherwise~\cite{arora2024simple}.

\begin{table}[H]
    \centering
    \begin{tabular}{ll}
         \textbf{Dataset name}& \textbf{Huggingface URL}  \\ \toprule
         E2E & \url{https://huggingface.co/datasets/e2e_nlg}\\
         CNN/DailyMail & \url{https://huggingface.co/datasets/cnn_dailymail}\\
         SQuAD & \url{https://huggingface.co/datasets/hazyresearch/based-squad}\\
         XSum & \url{https://huggingface.co/datasets/EdinburghNLP/xsum}\\
         TriviaQA & \url{https://huggingface.co/datasets/mandarjoshi/trivia_qa}\\
         Web NLG & \url{https://huggingface.co/datasets/web_nlg} \\ 
         Definition Extraction & \url{https://huggingface.co/datasets/nguha/legalbench}\\ \bottomrule
    \end{tabular}
    \caption{Datasets used.}
    \label{tab:huggingface_datasets_links}
\end{table}

Table \ref{tab:huggingface_model_links} contains links for all models used.

\begin{table}[H]
    \centering
    \begin{tabular}{ll}
         \textbf{Model name}& \textbf{Huggingface URL}  \\ \toprule
         Pythia-410M & \url{https://huggingface.co/EleutherAI/pythia-410m}\\
         Pythia-1B & \url{https://huggingface.co/EleutherAI/pythia-1b}\\
         Pythia-2.8B & \url{https://huggingface.co/EleutherAI/pythia-2.8b}\\
         Pythia-6.9B & \url{https://huggingface.co/EleutherAI/pythia-6.9b}\\
         Gemma-2B & \url{https://huggingface.co/google/gemma-2b-it}\\
         Incite-3B & \url{https://huggingface.co/togethercomputer/RedPajama-INCITE-Instruct-3B-v1}\\
         Dolly-3B & \url{https://huggingface.co/databricks/dolly-v2-3b}\\
         Llama-2-7B & \url{https://huggingface.co/meta-llama/Llama-2-7b-hf}\\
         Mistral-7B & \url{https://huggingface.co/mistralai/Mistral-7B-Instruct-v0.2}\\
         Vicuna-7B & \url{https://huggingface.co/lmsys/vicuna-7b-v1.5}\\
         Gemma-7B & \url{https://huggingface.co/google/gemma-7b}\\
         Nous Capybara & \url{https://huggingface.co/NousResearch/Nous-Capybara-7B-V1.9}\\
         Phi-2 & \url{https://huggingface.co/microsoft/phi-2}\\
         Llema-7B & \url{https://huggingface.co/EleutherAI/llemma_7b}\\ \bottomrule
    \end{tabular}
    \caption{Huggingface model URLs.}
    \label{tab:huggingface_model_links}
\end{table}

For the Alpaca leaderboard experiments, we run each trial by sampling 5 models from the following set of 10: \texttt{Nanbeige-Plus-Chat-v0.1}, \texttt{claude-2}, \texttt{Qwen1.5-110B-Chat}, \texttt{yi-large-preview}, \texttt{gemini-pro}, \texttt{Meta-Llama-3-70B-Instruct}, \texttt{Ein-70B-v0.1}, \texttt{mistral-large-2402}, \texttt{Storm-7B}, \texttt{FsfairX-Zephyr-Chat-v0.1}.

\subsection{Additional Results}

\begin{figure}[h!]
    \centering
    \includegraphics[scale=0.5]{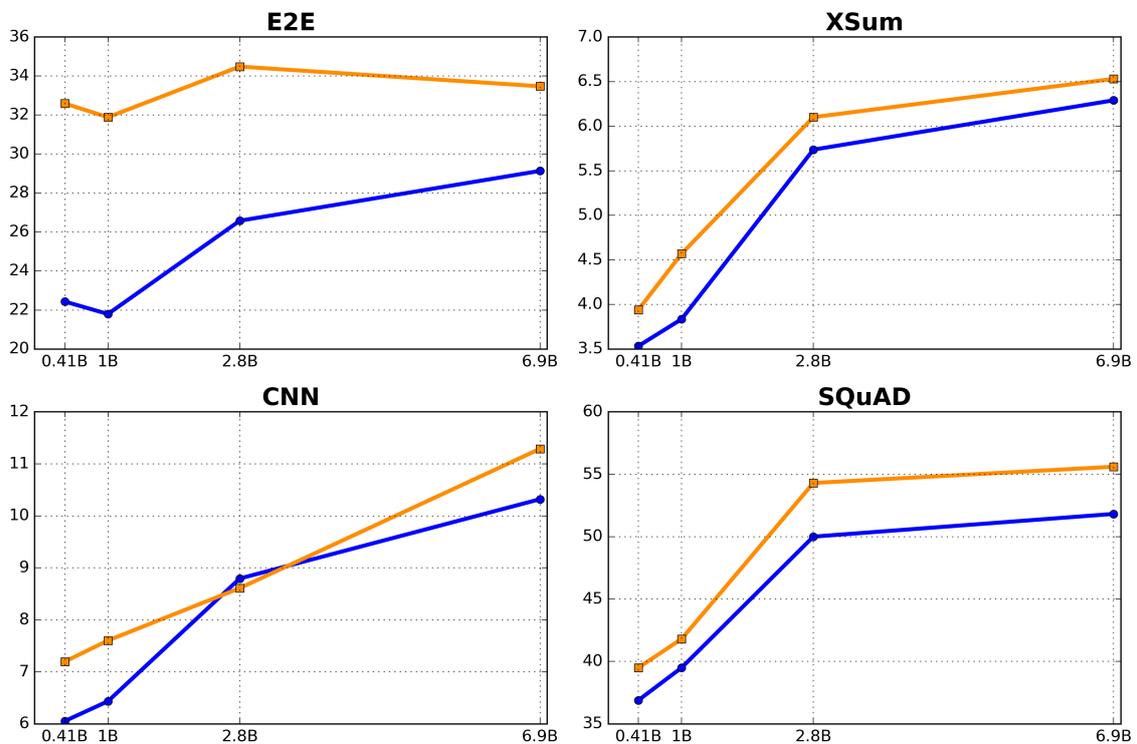}
    \caption{We compare \textsc{Random} (\textcolor{blue}{blue}) and \nameglobal (\textcolor{orange}{orange}) for prompt-selection on different sized models in the Pythia suite. The x-axis denotes model size, and the y-axis denotes performance (either rouge2 or accuracy).}
    \label{fig:scaling_results}
\end{figure}

\begin{table}[H]
\centering
\renewcommand{\arraystretch}{1.2}
\setlength{\tabcolsep}{7pt}
\begin{tabular}{@{}lcccccccc@{}}
\toprule
& & CNN & Def. Ext. & E2E & SQuAD & TriviaQA & WebNLG & XSum \\ \midrule
\multirow{3}{*}{3B}
& \small{\textsc{Random}} & 12.9 & 52.4 & 27.3 & 59.6 & \underline{32.7} & 23.4 & \underline{4.5} \\
& \small{\nameglobal} & \underline{\textbf{14.3}} & \underline{\textbf{61.5}} & \underline{\textbf{31.8}} & \underline{60.7} & 32.1 & \underline{\textbf{30.7}} & \underline{4.5} \\
& \small{\nameglobal-\textsc{Train}} & \textbf{14.3} & \textbf{61.5} & 24.7 & 60.7 & 32.1 & \textbf{30.7} & 4.5 \\
& \small{\textsc{Best-on-Val}} & 13.0 & 60.5 & 31.1 & \textbf{66.4} & \textbf{38.7} & 30.3 & \textbf{5.3} \\
\midrule
\multirow{3}{*}{7B}
& \small{\textsc{Random}} & 13.7 & 58.5 & 35.3 & 67.9 & 59.3 & 44.1 & 6.9 \\
& \small{\nameglobal} & \underline{\textbf{14.5}} & \underline{\textbf{70.9}} & \underline{\textbf{36.9}} & \underline{\textbf{76.2}} & \underline{\textbf{68.3}} & \underline{45.9} & \underline{\textbf{8.4}} \\
& \small{\nameglobal-\textsc{Train}} & \textbf{14.5} & \textbf{70.9} & 36.5 & \textbf{76.2} & \textbf{68.3} & 45.9 & \textbf{8.4} \\
& \small{\textsc{Best-on-Val}} & \textbf{14.5} & 69.4 & 36.7 & 74.0 & 65.8 & \textbf{48.3} & 8.3 \\
\bottomrule
\end{tabular}
\caption{We compare \nameglobal to \nameglobal-\textsc{Train}, for which weights are learned on a hold-out set. We provide results from baseline methods for reference. Underlined values are the best performing \textit{unsupervised} methods. Bold values are the best performing \textit{overall} methods. We report rouge2 scores for CNN, XSum, WebNLG, and E2E, and accuracy for the rest. All metrics are scaled to 0-100.}
\label{tab:smoothie-train-test-selection-comparison}
\end{table}

\begin{table}[H]
\centering
\renewcommand{\arraystretch}{1.3}
\setlength{\tabcolsep}{10pt}
\begin{tabular}{lccccc}
\toprule
& \multicolumn{2}{c}{\textbf{3B}} & \multicolumn{2}{c}{\textbf{7B}} \\
\cmidrule(lr){2-3} \cmidrule(l){4-5}
\textbf{Method} & \distacc & \distr & \distacc & \distr \\ \midrule
\small{\random} & 48.7 & 17.0 & 65.4 & 25.0 \\ \midrule
\small{\pairrm} & 53.9 & 19.0 & 71.8 & 25.5 \\ \midrule
\small{\lknn} & 51.0 & 16.8 & 71.7 & 26.2 \\ \midrule
\small{\bestensemble} & 52.3 & 18.1 & 73.2 & 26.4 \\ \midrule
\small{\nameglobal} & 51.3 & 18.1 & 66.5 & 26.1 \\ \midrule
\small{\namelocal} & \textbf{58.7} & \textbf{20.2} & \textbf{75.0} & \textbf{26.9} \\ \midrule
\small{\nameglobal-\textsc{train}} & 51.3 & 18.1 & 66.5 & 26.1 \\ \midrule
\small{\namelocal-\textsc{train}} & 50.7 & 18.8 & 70.9 & 26.0 \\ 
\bottomrule
\end{tabular}
\caption{We compare \namelocal to \namelocal-train, for which weights are learned on a hold-out set, on the 3B and 7B ensembles for multi-task distributions. \distacc and \distr are measured with accuracy and rouge2 respectively. Bold values indicate the best performing method for each dataset and model size. Metrics are scaled to 0-100. Other baseline methods are provided for comparison.}
\label{tab:smoothie-train-test-local-comparison}
\end{table}

\begin{table}[H]
\centering
\renewcommand{\arraystretch}{1.3}
\setlength{\tabcolsep}{10pt}
\begin{tabular}{lccccc}
\toprule
& \multicolumn{2}{c}{\textbf{3B}} & \multicolumn{2}{c}{\textbf{7B}} \\
\cmidrule(lr){2-3} \cmidrule(l){4-5}
\textbf{Method} & \distacc & \distr & \distacc & \distr \\ \midrule
\small{\random} & 48.7 & 17.0 & 65.4 & 25.0 \\ \midrule
\small{\pairrm} & 53.9 & 19.0 & 71.8 & 25.5 \\ \midrule
\small{\lknn} & 51.0 & 16.8 & 71.7 & 26.2 \\ \midrule
\small{\bestensemble} & 52.3 & 18.1 & 73.2 & 26.4 \\ \midrule
\small{\namelocal} (BGE-small~\cite{bge_embedding}) & \textbf{59.3} & 19.7 & 74.6 & \textbf{27.1} \\ \midrule
\small{\namelocal} (SBERT~\cite{reimers-2019-sentence-bert}) & 58.7 & \textbf{20.2} & \textbf{75.0} & 26.9 \\
\bottomrule
\end{tabular}
\caption{Comparing \namelocal with different embeddings on the 3B and 7B ensembles for multi-task distributions. \distacc and \distr are measured with accuracy and rouge2 respectively. Bold values indicate the best performing method for each dataset and model size. Metrics are scaled to 0-100.}
\label{tab:smoothie-routing-embedding-ablation}
\end{table}

\begin{figure}[H]
    \centering
    \includegraphics[width=0.7\linewidth]{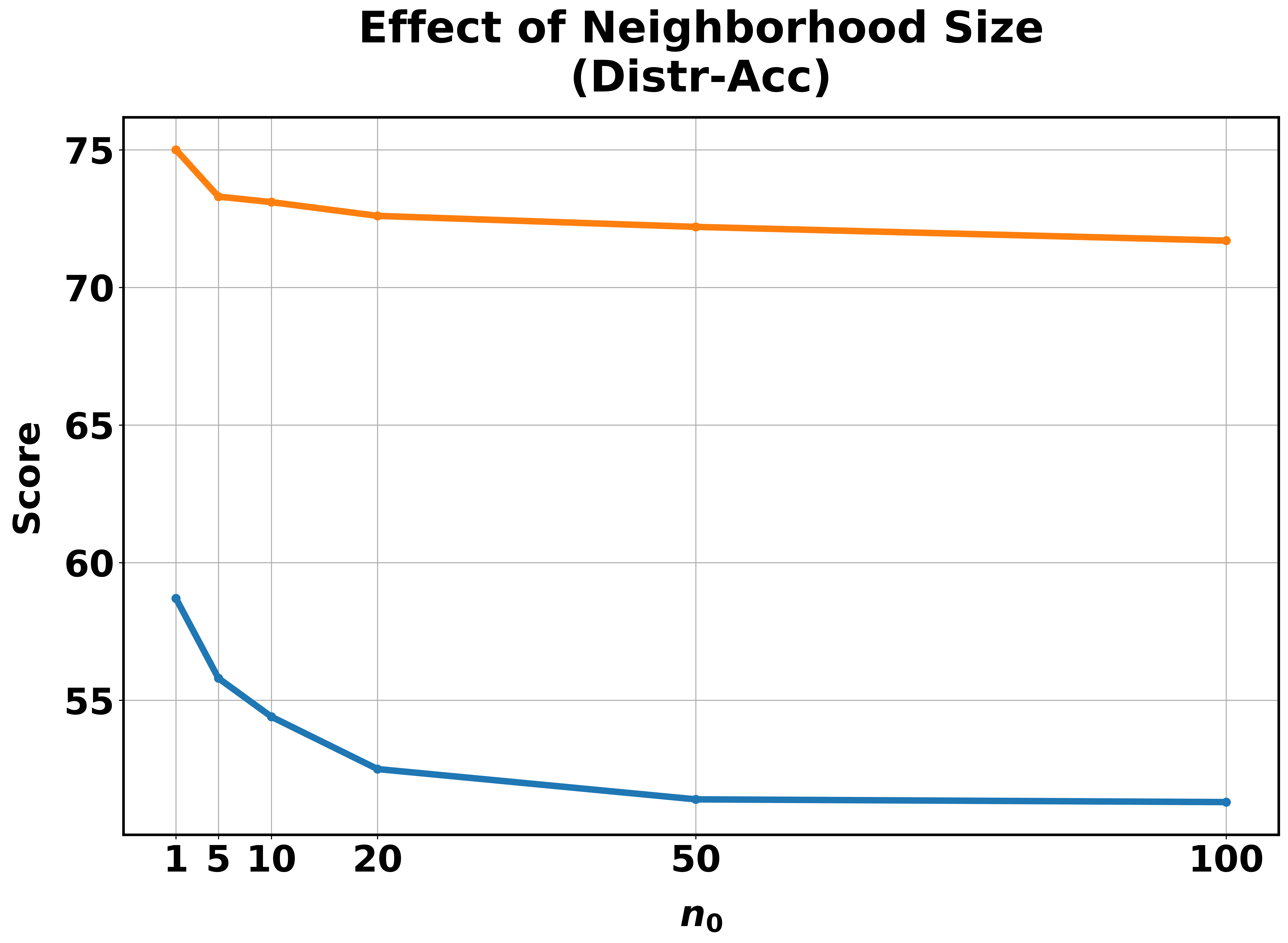}
    \caption{We measure how \namelocal's performance on \distacc changes as $n_0$ changes.}
    \label{fig:smoothie_n0_ablation_acc_group}
\end{figure}

\begin{figure}[H]
    \centering
    \includegraphics[width=0.7\linewidth]{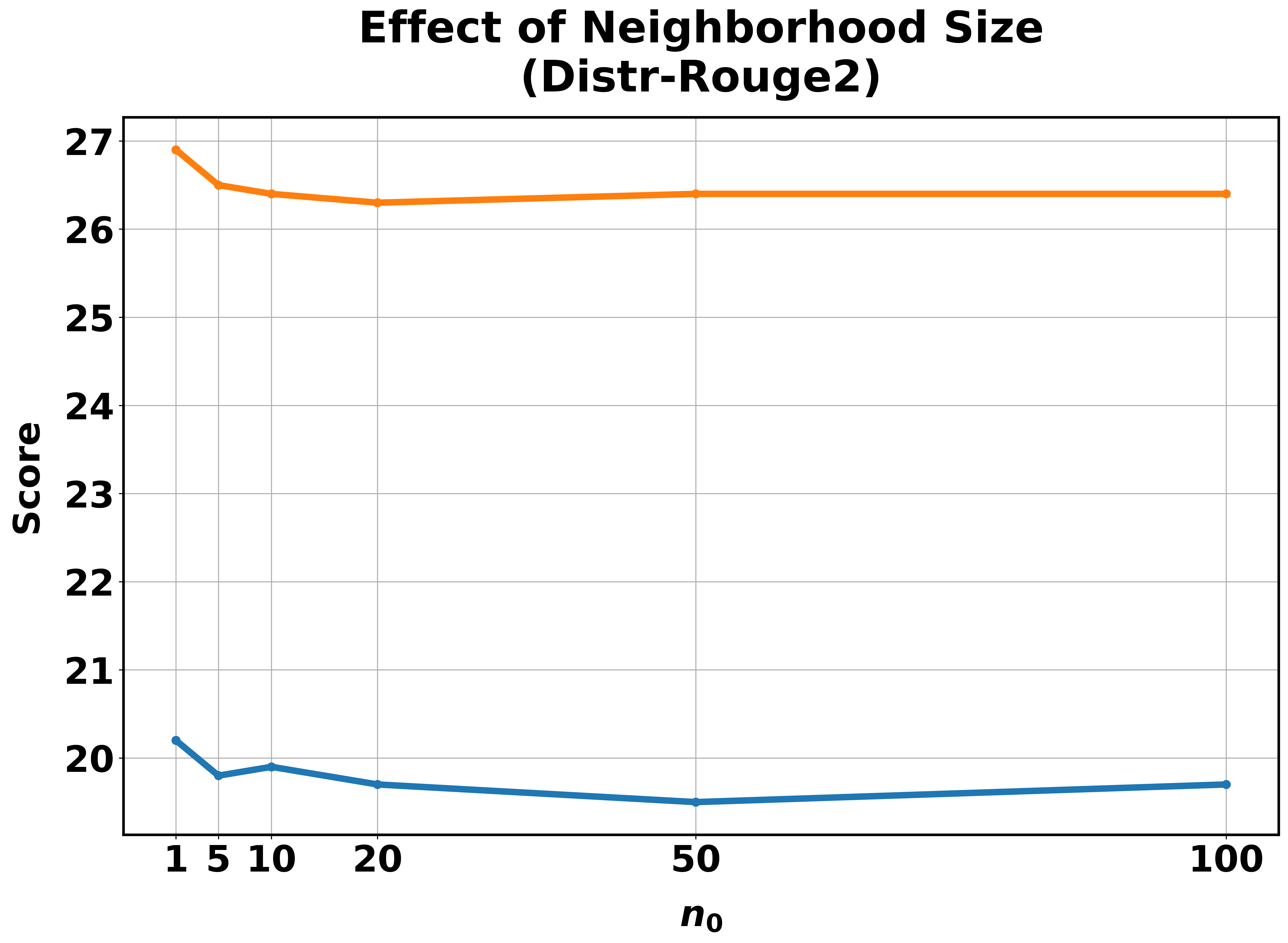}
    \caption{We measure how \namelocal's performance on \distr changes as $n_0$ changes.}
    \label{fig:smoothie_n0_ablation_rouge2_group}
\end{figure}

\begin{figure}[H]
    \centering
    \includegraphics[width=0.7\linewidth]{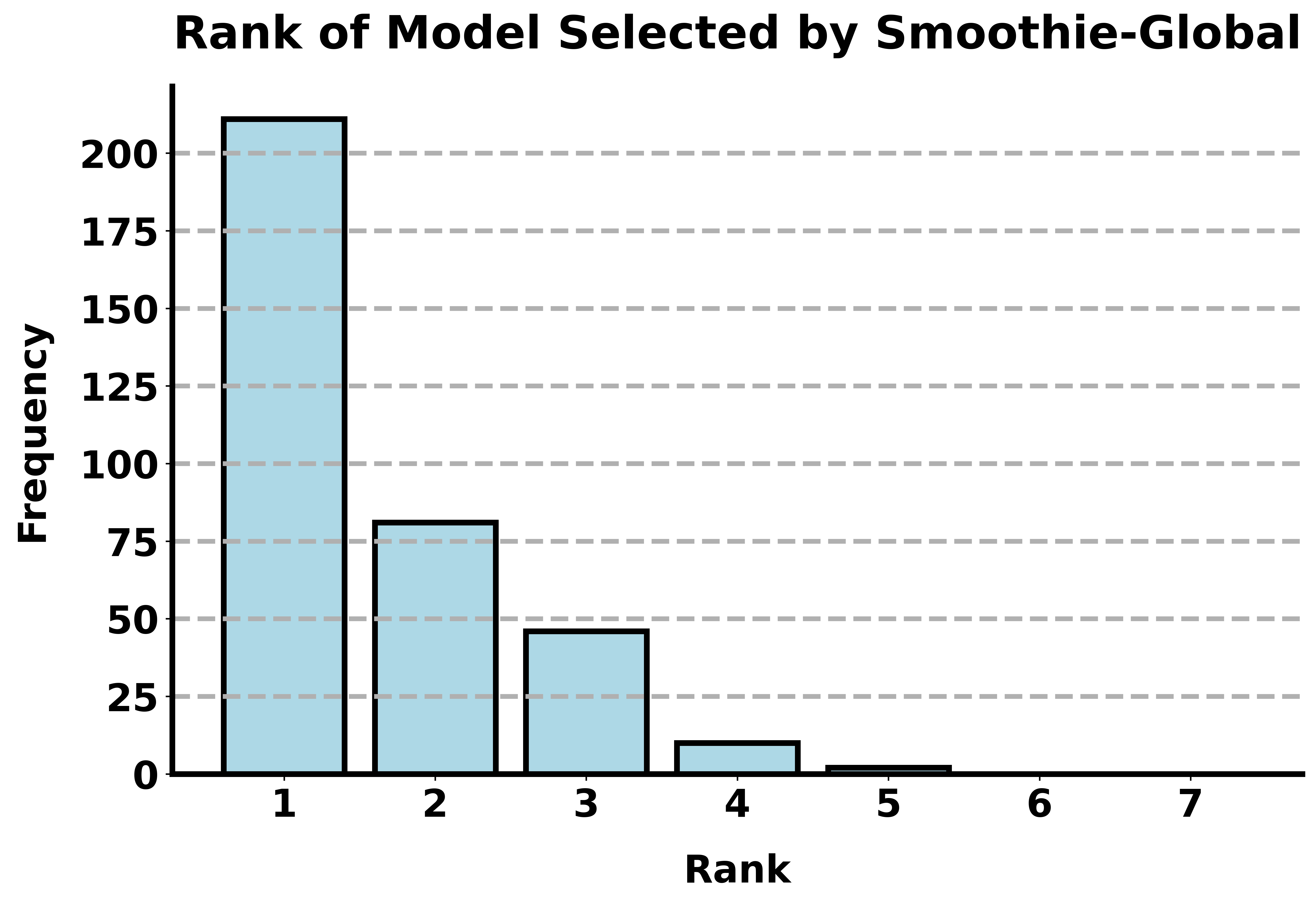}
    \caption{We evaluate \nameglobal's ability to identify the best model by randomly sampling 50 ensembles of size 4-7 LLMs from a pool of the LLMs contained in the 3B and 7B ensembles. We apply \nameglobal to select the best LLM from within each of these ensembles across the 7 NLG tasks, and measure the rank (relative to the ensemble) of the LLM selected by \nameglobal.}
    \label{fig:smoothie_many_ensemble_ablation}
\end{figure}

\begin{table}[H]
    \centering
    \begin{tabular}{cc}
        \toprule
        Method & ChatGPT-Rank ($\downarrow$) \\ \midrule
        \random & 5.96 \\ \midrule
        \nameglobal & 3.91 \\ \bottomrule
    \end{tabular}
    \caption{Results for \nameglobal and baselines on MixInstruct.}
    \label{tab:mix_instruct_selection_results}
\end{table}

\begin{table}[H]
    \centering
    \begin{tabular}{cc}
        \toprule
        Method & Accuracy  \\ \midrule
        \random & 28.4 \\ \midrule
        \bestonval & 37.5\\ \midrule
        \nameglobal & 37.5 \\ \bottomrule
    \end{tabular}
    \caption{Results for \nameglobal and baselines on GSM8K. We report accuracy, with scores scaled to 0-100.}
    \label{tab:gsm8k_selection_results}
\end{table}

\begin{figure}[H]
    \centering
    \includegraphics[width=0.7\linewidth]{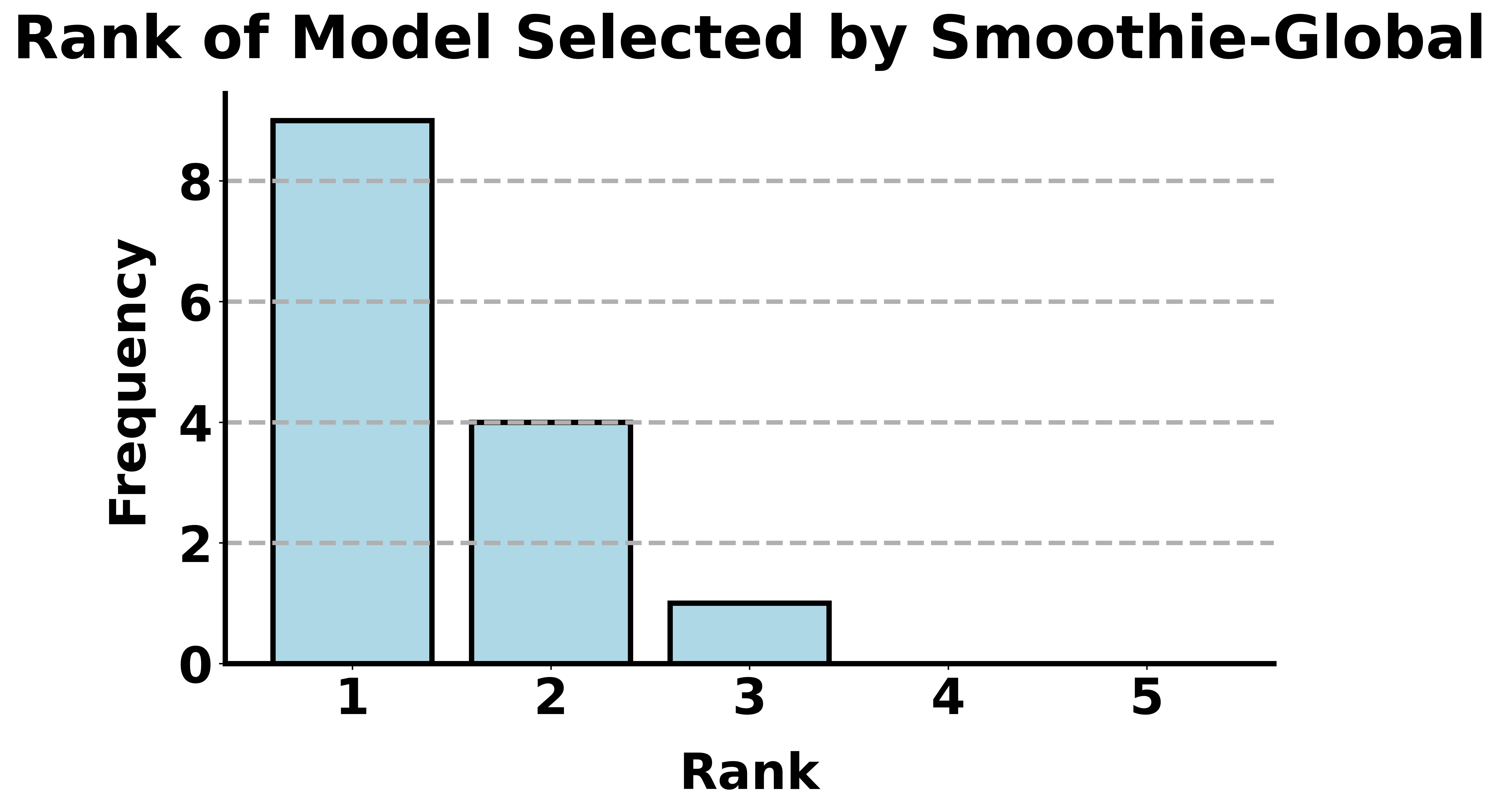}
    \caption{We construct a histogram over the rank of the LLM selected by \nameglobal across both the 3B and 7B ensembles, for 7 NLG tasks.}
    \label{fig:smoothie_nlg_ensemble_ranks}
\end{figure}

\end{document}